\documentclass{article}

\usepackage{microtype}
\usepackage{graphicx}
\usepackage{subfigure}
\usepackage{booktabs} % for professional tables

\usepackage{hyperref}

\usepackage[accepted]{icml2025}

\usepackage{amsmath}
\usepackage{amssymb}
\usepackage{mathtools}
\usepackage{amsthm}

\usepackage[capitalize,noabbrev]{cleveref}

\theoremstyle{plain}
\newtheorem{theorem}{Theorem}[section]

\theoremstyle{definition}

\newtheorem{assumption}[theorem]{Assumption}
\theoremstyle{remark}

\usepackage[textsize=tiny]{todonotes}

\newcommand{\E}{\mathbb{E}}
\newcommand{\var}{\textbf{var}}

\newcommand{\one}{\textbf{1}}

\usepackage[normalem]{ulem}

\definecolor{darkgreen}{rgb}{0.0, 0.5, 0.0}
\newcommand{\red}[1]{\color{red}#1}
\newcommand{\green}[1]{\color{green!50!black}#1}

\icmltitlerunning{Improving realistic long-tailed semi-supervised learning with doubly robust estimation}

\begin{document}

\twocolumn[
\icmltitle{Improving realistic semi-supervised learning with doubly robust estimation}


\begin{icmlauthorlist}
\icmlauthor{Khiem Pham}{cornell}
\icmlauthor{Charles Herrmann}{deepmind}
\icmlauthor{Ramin Zabih}{cornell}
\end{icmlauthorlist}

\icmlaffiliation{cornell}{Cornell Tech}
\icmlaffiliation{deepmind}{Google DeepMind}

\icmlcorrespondingauthor{Khiem Pham}{dkp45@cornell.edu}

\icmlkeywords{Machine Learning, ICML}

\vskip 0.3in
]

\printAffiliationsAndNotice{}

\begin{abstract}
% One of the major challenges in semi-supervised learning is label shift, where the unlabeled class distribution is different from the labeled class distribution and is also unknown. This creates difficulty for the standard pseudo-labeling approach because the classifiers do not transfer well from labeled to unlabeled data. As a result it is typically assumed that the unlabeled data class distribution is either known apriori, or estimated on-the-fly using the pseudo-labels themselves. Based on the intuition that learning the average of class conditionals is easier than learning them point-wise, we show that by using better methods, the unlabeled class distribution can be estimated accurately, which then results in better pseudo-labels and eventually improves point-wise classification. We show this improvement in the recently proposed Simpro.

A major challenge in Semi-Supervised Learning (SSL) is the limited information available about the class distribution in the unlabeled data.
In many real-world applications this arises from the prevalence of long-tailed distributions, where the standard pseudo-label approach to SSL is biased towards the labeled class distribution and thus performs poorly on unlabeled data.
Existing methods typically assume that the unlabeled class distribution is either known a priori, which is unrealistic in most situations, or estimate it on-the-fly using the pseudo-labels themselves. 
We propose to explicitly estimate the unlabeled class distribution, which is a finite-dimensional parameter, \emph{as an initial step}, using a doubly robust estimator with a strong theoretical guarantee; this estimate can then be integrated into existing methods to pseudo-label the unlabeled data during training more accurately.
Experimental results demonstrate that incorporating our techniques into common pseudo-labeling approaches improves their performance. 

\end{abstract}
\section{Introduction}
\label{sec:intro}

\noindent
Semi-supervised learning (SSL) aims to augment the small labeled set of data with a large unlabeled set of data \cite{chapelle2009semi}. 
This is of considerable practical significance since in many applications unlabeled data is easily available but the labeling effort is very costly. 
Many semi-supervised learning methods have proven successful, even given very small amounts of labeled data. 
However there is very limited information about the unlabeled data, since we only have access to their features and not the unlabeled class distribution.
We will write the labeled and unlabeled class distributions as $P(Y|A=1)$ and $P(Y|A=0)$, respectively.
In some situations $P(Y|A=0)$ is known a priori, but in many practical applications it is unknown and difficult to estimate from $P(Y|A=1)$.
In particular, the distribution $P(Y|A=0)$ is frequently long-tailed.
%, and can evolve over time as new data is collected.
This SSL variant, where the unlabeled class distribution $P(Y|A=0)$ is unknown and differs from $P(Y|A=1)$, is sometimes called \textit{realistic long-tailed semi-supervised learning} (RTSSL).
This topic has drawn considerable recent interest (see for example \cite{simpro,kim2020distribution,crest,oh2022daso,acr,cpe}) since it reflects realistic assumptions in many applications.

\begin{figure*}[th]
\begin{center}
\includegraphics[width=0.7\textwidth]{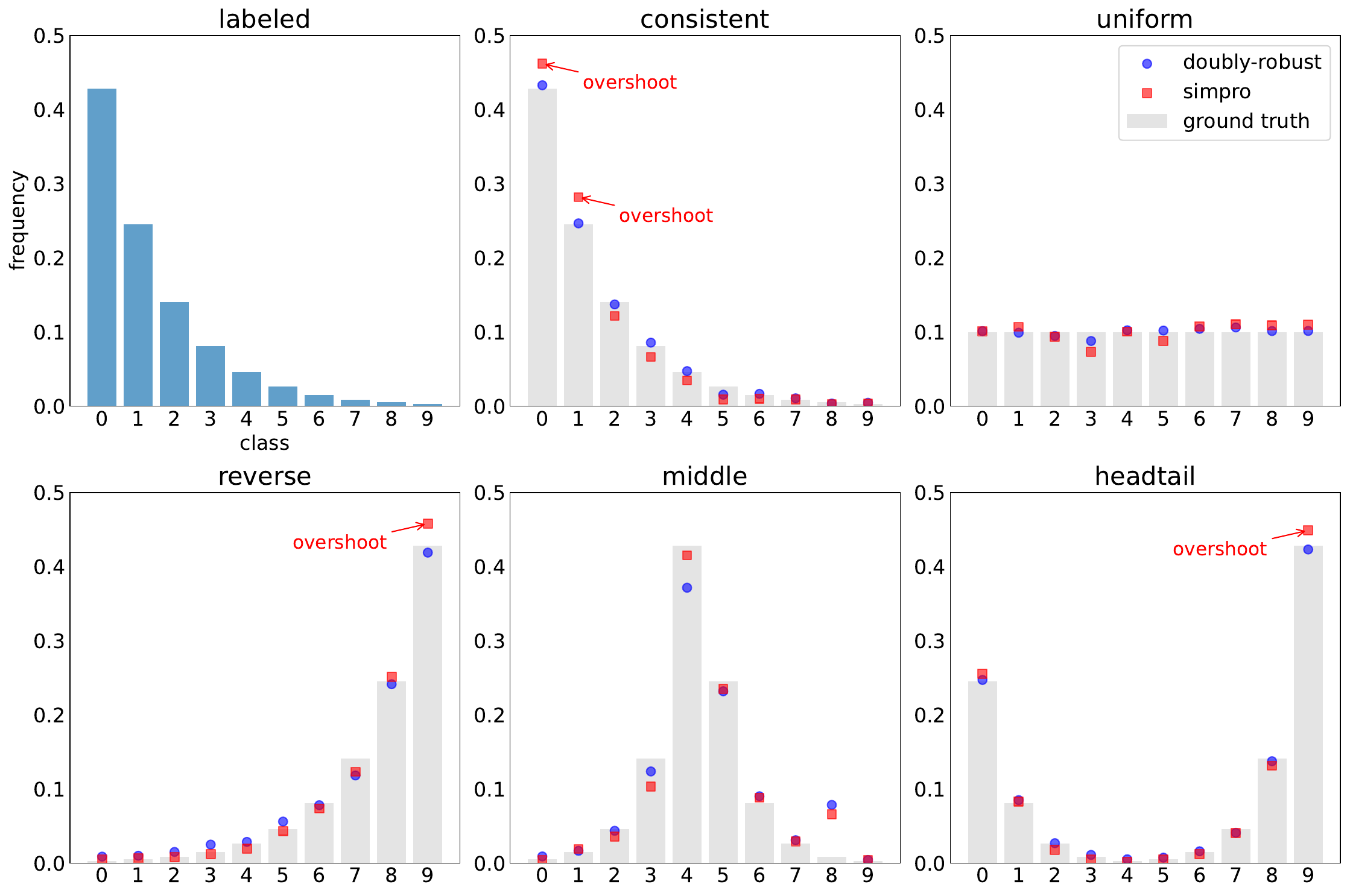}
\end{center}
\caption{The labeled class distribution and 5 possible unlabeled class distributions studied in \cite{simpro}. SimPro significantly overestimates the head classes in consistent, reverse and head-tail settings. Our doubly-robust estimate is more accurate at the head classes as well as the overall distribution in all but the middle setting, as measured in total variation distance in \cref{tab:cifar10-tv}. Our proposed 2-stage SimPro+ outperforms SimPro in classification accuracy in the middle setting as well.}
\label{fig:distribution}
\end{figure*}
% In this paper, we explicitly estimate the unlabeled class distribution $P(Y|A=0)$ as an initial first step. We note that a good approximation of this distribution can lead to good classification accuracy and vice versa, but a good classifier is not necessary to obtain a good class distribution because the former is infinite dimensional while the latter is not. We apply non-ignorable missingness, where the probability of data being missing depends on its value; this is an important statistical problem formulation that appears to have been overlooked in the vision community. 
In this paper, we explicitly estimate the unlabeled class distribution $P(Y|A=0)$ as a separate first step. We note that existing methods that estimate and use this distribution during training produce biased estimate. In particular, SimPro \cite{simpro} tends to significantly overestimate the head classes as shown in \cref{fig:distribution} in 4 out of 5 unlabeled class distributions studied. In contrast, our proposed doubly-robust estimator are more accurate. Our technique derives from semi-parametric efficiency theory predominantly studied in causal inference and has well-understood and strong theoretical guarantee \cite{dml}. Leveraging this improvement, we plug this first-stage estimate into a second stage algorithm for training the final classifier.

We also adapt a maximum likelihood framework for semi-supervised learning with label shift. The framework uses and estimates a \textit{missingness mechanism} which encodes the tendency of a label to be in the labeled $(A=1)$ or unlabeled $(A=0)$ set, and allows learning from both sets from one missing-data likelihood, which we address with an Expectation-Maximization (EM) algorithm. 
The basic idea dates as far back as \cite{ibrahim1996parameter}, yet despite its simplicity, it is often overlooked in the label shift and vision community. We show that it naturally generalizes and extends FixMatch \cite{fixmatch}. The recent work of \cite{simpro} can be seen as the same algorithm but with different parameterization (see \cref{subsec:simpro}). 

\begin{figure*}[t]
\begin{center}
\includegraphics[width=0.7\textwidth]{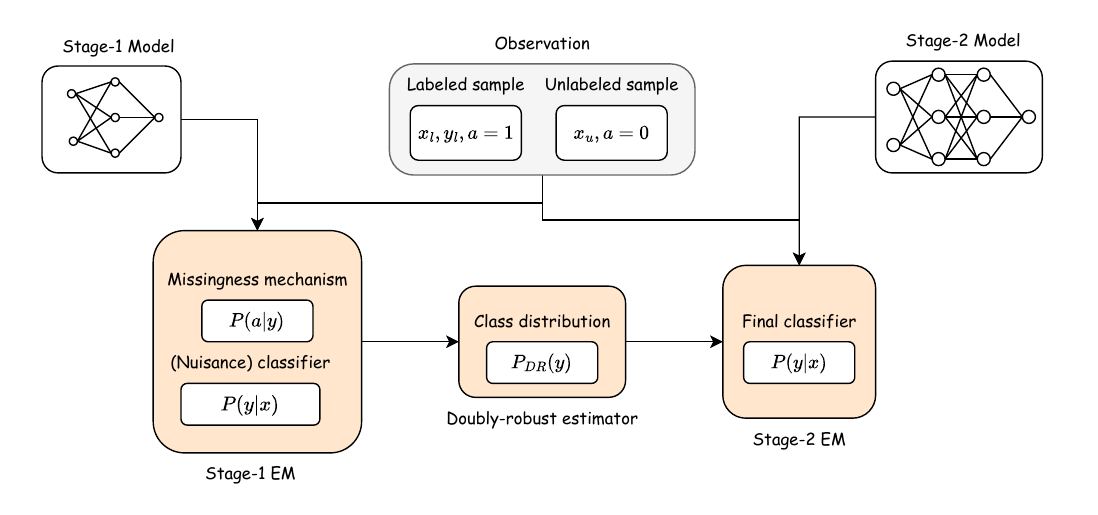}
\end{center}
\caption{Overview of our 2-stage method (\cref{subsec:2-stage}). In stage 1, we use Expectation-Maximization (EM, \cref{subsec:em}) to estimate the missingness mechanism and classifier from observable data. These quantities are used as nuisance components for the doubly-robust estimator of the class distribution \cref{eq:dr}. In stage 2, we can use EM or other existing methods that also use logit-adjustment with the (unlabeled) class distribution to estimate the final classifier. We use SimPro as our implementation of EM (\cref{subsec:simpro}). The network in stage 1 can be of equal or smaller size than the network in stage 2 (\cref{subsec:label}).}
\label{fig:algorithm}
\end{figure*}

In summary, we propose a 2-stage algorithm for RTSSL (see \cref{fig:algorithm}). The first stage uses maximum likelihood and EM to initially learn about the data. The first-stage estimates are used for a meta doubly-robust estimator which significantly improves on the initial estimate of $P(Y|A=0)$.
We then plug this estimate into existing pseudo-labeling-based techniques to learn the final classifier.
Experiments demonstrate that our method produces a significantly better estimate of $P(Y|A=0)$. 
We also show that we improve overall accuracy when our estimate is plugged into existing pseudo-labeling-based techniques.
Additional experimental results and some technical details are deferred to the supplemental.

\section{Background and Related Work}
\label{sec:background}

\textbf{Notation} We write the random variable $X \in \mathcal{X}$ for the feature(s) and  $Y \in \{1,\dots, C\}$ for the class among $C$ possible classes. We are given a labeled dataset $D_l = \{x_i,y_i\}_{i=1}^{N_l}$ and an unlabeled dataset $D_u = \{x_i\}_{i=N_l+1}^N$, where $x_i$ and $y_i$ are realizations of $X$ and $Y$. The training dataset is $D_t = D_l \cup D_u$. 
The auxiliary variable $A$ takes binary values and selects between the different class distributions $P(Y|A)$; let $A=1$ if the datapoint is in the labeled set and $A=0$ in the unlabeled set. 
Therefore $P(Y|A=0)$ is the class distribution the unlabeled set. The \textit{combined} class distribution $P(Y) = P(A=0)P(Y|A=0) + P(A=1)P(Y|A=1)$ is the class distribution of the combined dataset. For convenience, we also denote $P(Y|\textit{uniform})=1/C$ everywhere to be the uniform class distribution, noting that it is not another value of $A$. We assume that the class distribution of the test set is uniform throughout this paper.

\paragraph{Long-tailed Semi-supervised learning} is the intersection between long-tailed learning \cite{longtailedlearning} and semi-supervised learning \cite{chapelle2009semi}, and attempts to deal with two key real world problems: class distribution in the wild is often long-tailed with many classes having few samples; and the unlabeled data dwarfs the labeled data because of the advent of the web and the significant cost of large-scale manual labeling efforts. Pseudo labeling \cite{pseudolabeling, mixmatch, google-selftraining, temporalensembling} has become one of the prominent approaches in semi-supervised learning, and has been extended to the long-tailed case \cite{crest, abc}, although the unlabeled class distribution was assumed to be the same as the labeled class distribution \cite{remixmatch}. More recent work has tackled the unknown distribution case \cite{dcssl, rda, onnonrandommissinglabels, prg4ssl, acr, simpro, cpe, boat}.
% Pseudo labeling in the balanced case is also connected to semi-supervised EM \cite{entropyminimization}, which naturally raises the question whether a similar connection exists for the different and unknown unlabeled distribution case. 

\paragraph{(Balanced) Pseudo-labeling.}
Semi-supervised learning methods use a regularization loss on the unlabeled data in addition to the classification loss on the labeled data. A simple technique is to use the model's own predictions on the unlabeled data. Specifically, FixMatch \cite{fixmatch} keeps the maximum predictions that also fall above a certain confidence threshold and convert them into one-hot labels (operator $\delta$), which is called a pseudo label. For example, given a confidence threshold of $0.8$, a binary prediction $[0.1, 0.9]$ will be mapped to $[0, 1]$ while $[0.4, 0.6]$ to $[0, 0]$ under the operator $\delta$. FixMatch then minimizes the cross entropy loss between a strongly augmented version  and the pseudo label of a weakly augmented version  of the same unlabeled image:
\begin{equation}
L_u = - \sum_{i=N_l+1}^{N} \sum_{c=1}^C \delta (P(Y|\alpha(x_i)))_c \log P(Y=c|G(x_i))
\label{eq:fixmatch-unlabeled-loss}
\end{equation}
where $c$ is the class, $G$ is the strong augmentation, and $\alpha$ is the weak augmentation. FixMatch is simple and performant. However, it suffers when labeled and unlabeled class distributions are different, which label shift approach tries to address.

\paragraph{Label shift and logit-adjustment.} Label shift assumes that the probability of $X$ given $Y$ is unchanged:
\begin{equation}
P(X|Y,A) = P(X|Y)
\label{eq:label-shift}
\end{equation} 
i.e. feature $X$ is conditionally independent of the variable $A$. The posterior change in $P(Y|X,A)$ results from the difference between the class distributions i.e. $P(Y|A=0) \neq P(Y|A=1)$. 
If the class distributions are known, logit adjustment can be used to convert a classifier of one class distribution to another.
When label shift occurs between two datasets, classifiers performing well on one dataset may not perform well on the other.
For example, to adapt the labeled class distribution $P(Y|A=1)$ to the test class distribution $P(Y|\textit{uniform})$, we can use Bayes formula to get:
\begin{equation}
P(Y|X,\textit{uniform}) \propto P(Y|X,A=1) \frac{P(Y|\textit{uniform})}{P(Y|A=1)}
\label{eq:logit-adjustment}
\end{equation}
which is the basis of the post-hoc logit adjustment formula for long-tailed learning.

Label shift is the natural assumption in imbalanced / long-tailed learning where the target distribution is the uniform test distribution. Logit adjustment \cite{logitadjustment}, implicitly using this assumption, relies on the formula \cref{eq:logit-adjustment} to correct label shift in long-tailed data. When the test distribution is unknown, label shift adaptation methods exist that can estimate the unknown test distribution \cite{mlls, mllsishardtobeat, bbse, rlls} when given a good classifier of the source data. It is possible therefore to train on the labeled set and use a label shift adaptation method to estimate the unlabeled class distribution. This procedure is best suited to label shift test-time adaptation \cite{ttsla, minh-gpa} where the unlabeled data is not available during model training. In contrast, when additional unlabeled data is available, semi-supervised EM gives significantly better class distribution estimation.

\paragraph{Non-ignorable missingness.} This is a variant of missing data problems where the missing indicator $A$ can depend on both feature $X$ and outcome $Y$ \cite{rubin-missingdata}. 
The dependence on $Y$ distinguishes this variant from the standard ignorable missingness (missing at random) assumption \cite{tsiatis-missingdata}. The label shift assumption \cref{eq:label-shift} further assumes that only $Y$ causes $A$, and this assumption is sufficient to \textit{identify} the true data distribution, meaning that no two distributions can generate our missing data \cite{labelshift-nonignorable, arelabelsinformative}.

\paragraph{Doubly robust (DR) estimation} 
\label{subsec:related-work-dr}
This approach has roots in semi-parametric efficiency theory \cite{kennedy-dr, dml}. The most successful application of DR is the estimation of the average treatment effect in causal inference \cite{tsiatis-missingdata, pham2023stable}, which is an example of ignorable missingness. Recently doubly machine learning \cite{dml, riesznet} takes double robustness further by showing that powerful machine learning methods such as neural networks can be used to deal with high-dimensional and complex data while at the same time making valid inference about the target statistics. The applications of DR in modeling more complex data than traditionally studied in statistics have recently gained significant interest \cite{dragonnet, riesznet, zhang2023towards}. We contribute to this line of work, but furthermore shows that we can plug in this estimation to improve the final classification itself.

Our work builds and improves on \cite{simpro}. Specifically, we show in \cref{subsec:simpro} that it is a reparameterization of the semi-supervised EM algorithm in \cref{subsec:em}, and we use it as the training method for both stages of our algorithm.
Our work is also close to \cite{arelabelsinformative, onnonrandommissinglabels} who also note the connection to non-ignorable missingness and propose doubly robust estimation of the loss.
This loss remains consistent even when the pseudo labels are arbitrarily bad, in a similar spirit to \cite{schmutz-drloss, drst}, as long as the missingness mechanism is correct. Thus they try to safeguard against wrong un-adjusted labels. We on the other hand try to improve the label's quality via EM and adjustment by the doubly robust estimation of the unlabeled class distribution.
An important weaknesses of the doubly robust loss \cite{arelabelsinformative} is that it involves inverse-weighting \cite{cui-effective} which is prone to unstable training \cite{balanced-meta-softmax}. We provide more detail about the doubly-robust risk in \cref{subsec:dr-risk}, and experimentally compare with it in \cref{subsec:ablation-1}
\section{Our approach}
\label{sec:method}
\subsection{Label shift Expectation Maximization}
\label{subsec:em}
When pseudo-labeling is applied naively, a classifier trained on the labeled set with class distribution $P(Y|A=1)$ may not do well on the unlabeled set that has a different class distribution $P(Y|A=0)$ thus resulting in incorrect pseudo labels for training and consequently confirmation bias \cite{confirmationbias}. We can not straightforwardly adapt to the unlabeled class distribution because it is unknown. In the following, we detail a likelihood maximization framework that eventually is shown to generalize pseudo-labeling to the label shift case. Using the indicator $A$, we can write the observed (or missing) data log-likelihood as
\begin{equation}
\begin{aligned}
L(\theta) &= \sum_{i=1}^{N_l} \log P(X=x_i,Y=y_i, A=1 | \theta)\\
&+ \sum_{i=N_l+1}^N \log P(X=x_i, A=0 | \theta),
\label{eq:likelihood}
\end{aligned}
\end{equation}
%\vspace*{-.05in}
where $\theta$ represents the parameter of the joint distribution $P(X,A,Y)$. This likelihood consists of the labeled term and an unlabeled term with a missing $Y$. Immediately, we can maximize $L(\theta)$ by writing the unlabeled term as a $Y$-marginalization of the joint as in \cite{arelabelsinformative}. As we will use EM to maximize $L(\theta)$, we apply Jensen inequality to the each term in the second sum using the posterior weight $\omega^t(x,y) = P(Y=y|X=x,A=0,\theta^t)$ where $\theta^t$ is value of $\theta$ in previous EM iteration, to get the lower bound
\vspace*{-.15in}
\begin{equation}
\begin{aligned}
&Q(\theta|\theta^t) = \sum_{i=1}^{N_l} \log P(X=x_i, Y=y_i, A=1 | \theta)\\
&+ \sum_{i=N_l+1}^N \sum_{c=1}^C \omega^t(x_i,c) \log P(X=x_i, Y=c, A=0 | \theta)
\label{eq:e-step}
\end{aligned}
\end{equation}
This is the E-step of EM and we have found the "pseudo-label" $\omega^t(x,c)$ for our unlabeled data, reducing the problem to a supervised learning one for the moment. Now we need to decide how to decompose the joint $P(X,Y,A|\theta)$ which decides what the parameter specification will be. It is natural that we use the invariance $P(X|Y,A) = P(X|Y)$ in \cref{eq:label-shift} to decompose $P(X|Y)P(Y|A)P(A)$, but this requires generative modeling for $P(X|Y)$. Instead, we use $P(A|Y)P(Y|X)P(X)$, which means we only need to learn a classifier $P(Y|X)$ and a finite-dimensional $P(A|Y)$, which are recipes for the posterior weight $\omega^t(x,y)$. With this, we get
\vspace*{-.2in}
\begin{equation}
\begin{aligned}
% Q(\theta|\theta^t) &= \sum_{i=1}^{N_l} \left[\log P(y_i | x_i, \theta) + \log P(a_i=1|y_i, \theta)\right] \\
% &+ \sum_{i=N_l+1}^{N} \sum_{c=1}^C P(y_i=c|x_i,a_i=0,\theta^t)\left[\log P(y_i=c|x_i, \theta) + \log P(a_i=0|y_i=c, \theta)\right]\\
Q(\theta|\theta^t) &= \sum_{i=1}^N \sum_{c=1}^C \gamma_i(c) \log P(Y=c|X=x_i,\theta)\\ 
&+ \sum_{c=1}^C \sum_{a=0}^1 \zeta_c(a) \log P(A=a|Y=c,\theta)
\label{eq:m-step}
\end{aligned}
\end{equation}
where $\gamma_i(c) = \one(y_i=c)$ for $i \le N_l$ and $P(Y=c|X=x_i,A=0,\theta^t)$ for $i > N_l$. $\zeta_c(1) = \sum_{i=1}^{N_l} \one(y_i=c)$ and $\zeta_c(0) = \sum_{i=N_l+1}^N P(Y=c|X=x_i,A=0,\theta^t)$. This means that maximizing $L(\theta|\theta^t)$ is equivalent to minimizing a sum of cross entropy losses. To compute the posterior weight $\omega^t(x,c)$, we use Bayes law:
\vspace*{-.1in}
\begin{equation}
\omega^t(x,c) \propto P(Y=c|X=x, \theta^t)P(A=0|Y=c, \theta^t)
\label{eq:e-step-posterior}
\end{equation}

\noindent In summary, the 2 steps of the EM are:

\medskip
\noindent
\textbf{E-step:} Given $P(Y=y|X=x,\theta^t)$ and $P(A=0|Y=y,\theta^t)$, set $\omega^t(x,c)$ according to \cref{eq:e-step-posterior}

\noindent
\textbf{M-step:} Given $\omega^t(x,c)$, find the new $P(Y|X,\theta)$ and $P(A|Y,\theta)$ by maximizing $Q(\theta|\theta^t)$.

\subsection{Label-shift Fixmatch and SimPro}
\label{subsec:simpro}
Pseudo labeling methods such as Fixmatch has a deep connection with Expectation-Maximization. Indeed, \cref{eq:fixmatch-unlabeled-loss} without the 3 operators is just the unlabeled term in \cref{eq:e-step} and $P(Y|A=1) = P(Y|A=0) = P(Y|\textit{uniform})$. SimPro is a recent work which also derives an almost equivalent EM formula to ours. They used a similar E-step but also applied Fixmatch's confidence thresholding and augmentation. Their M-step parameterizes the distribution as 2 parameters $\frac{P(X|Y)}{P(X)}$ and $P(Y|A=0)$. This is just another decomposition of the unlabeled log-likelihood term in \cref{eq:e-step} up to a constant $P(A=0)$:
\begin{equation}
%\begin{aligned}
\frac{P(X|Y)}{P(X)} P(Y|A=0) 
\propto \frac{P(Y|X)}{P(Y)}P(Y) P(A=0|Y)
%\end{aligned}
\label{eq:equi-simpro}
\end{equation}
Instead of canceling out $P(Y)$, however, SimPro uses a logit adjustment loss \cite{logitadjustment} for the first term in \cref{eq:m-step}:
\vspace*{-.1in}
\begin{equation}
\begin{aligned}
- \sum_{i=1}^N \sum_{c=1}^C \gamma_i(c) \log &\Big\{P(Y=c|X=x_i,\textit{uniform},\theta) \\
&+ P(Y=c)\Big\}
\end{aligned}
\label{eq:simpro-la-loss}
\end{equation}
%\vspace*{-.1in}
As $P(Y=c)$ is unknown, they use its running estimate. The model is automatically logit adjusted to the uniform test distribution during training. In contrast, if the model is $P(Y|X)$ in \cref{eq:m-step}, we can apply post-hoc logit adjustment. As shown in \cite{logitadjustment}, the logit adjustment loss is often slightly better, and this is what we find experimentally as well. Other than this difference, we can recover the class distribution $P(Y|A)$ from the missingness mechanism $P(A|Y)$ and because $P(A)$ is known, so they are learned equivalently.

\subsection{Our 2-stage algorithm}
\label{subsec:2-stage}
% We have shown that we can generalize Fixmatch's pseudo labeling to the label shift case by applying the logit adjustment to the pseudo label in \cref{eq:e-step-posterior}. This adjustment term, the missingness mechanism $P(A|Y)$, can be learned at the same time by minimizing the second term in \cref{eq:m-step}. Since $P(Y|A)$ or equivalently $P(A|Y)$ is a finite-dimensional parameter, it can be much easier to estimate than the classification task $P(Y|X)$ and can be done first. Therefore, we propose to separate the estimation of the unlabeled class distribution from the classifier learning. In the first stage, we strive to achieve a good estimation of $P(Y|A=0)$. Then, we freeze this parameter and plug it into EM or SimPro, or any other approaches that also uses distribution alignment \cite{remixmatch}.

Figure~\ref{fig:algorithm} shows the overview of our algorithm. During training, we use the current model's predictions and adjust it to the unlabeled class distribution $P(Y|A=0)$. The quality our first-stage estimation of $P(Y|A=0)$ has a direct impact on the pseudo label accuracy, as highlighted in Theorem 3.1 of \cite{lsc}. Briefly, the error gap between the adjusted model and the Bayes-optimal model can be bounded by the sum of an error term induced by the model's performance on the training data and another error term induced by the quality of our unlabeled distribution estimation. Therefore, we should aim for the highest estimation quality we can get in the first stage. To this end, we present 3 possible estimators for the combined class distribution $P(Y)$, the outcome regression (OR) estimator, inverse probability weighted (IPW) estimator and the doubly robust (DR) estimator. The unlabeled class distribution $P(Y|A=0)$ can be recovered by noting that $P(Y) = \sum_a P(A=a) P(Y|A=a)$ and that $P(A)$ and the labeled class distribution $P(Y|A=1)$ is known. The OR estimator is simply the average of the model's predictions
% \vspace*{-.1in}
\begin{equation}
\Psi_{or}(c) = \frac{1}{N}\sum_{i=1}^N P(Y=c|X=x_i,\theta)
\end{equation}
where the summation takes both labeled and unlabeled data. 

Another estimator is the inverse probability weighted (IPW) estimator. Suppose that we have the ground truth missingness mechanism $P(A|Y)$, then we have the following identity:
\begin{equation}
P(Y=c) = \E_O\left[\frac{\one(A=1)}{P(A=1|Y)} \one(Y=c)\right]
\end{equation}
where $O$ is a random variable representing one observation from the combined dataset, which is complete ($O=(X,A=1,Y)$) if the datapoint is from the labeled set and missing $(X,A=0)$ if unlabeled set. The indicator $\one(A=1)$ means that we are not actually using ground truth labels from the unlabeled set, but up-weighting the existing labels from the labeled set by the missingness mechanism. Replacing expectation with sample average and $P(A=1|Y)$ with an estimation $P(A=1|Y,\theta)$, we get our IPW estimator of $P(Y)$, which depends on $\theta$
% \vspace*{-.15in}
\begin{equation}
\Psi_{ipw}(\theta)(c) = \frac{1}{N}\sum_{i=1}^N \frac{\one(a_i=1)}{P(A=1|Y=y_i,\theta)}\one(y_i=c)
\end{equation}

\noindent
\vspace*{-.3in}
\paragraph{Our doubly robust estimator}
It is worth noting that each estimator above (OR or IPW) uses only one part of the distribution, either $P(Y|X)$ or $P(A|Y)$. The DR estimator takes advantage of both of these quantities. It is
\vspace*{-.1in}
\begin{equation}
\label{eq:dr}
\begin{aligned}
&\Psi_{dr}(\theta)(c) = \frac{1}{N}\sum_{i=1}^N\Bigg[ P(Y=c|X=x_i,\theta) + \\
&\frac{\one(a_i=1)}{P(A=1|Y=y_i,\theta)}( \one(y_i=c) - P(Y=c|X=x_i,\theta) )\Bigg]
\end{aligned}
\end{equation}
$\Psi_{dr}(\theta)$ is called doubly-robust because, given either a correct $P(Y|X)$ or $P(A=1|Y)$, we will get an unbiased estimate of $P(Y)$. 
We need to learn both of these quantities from finite data which means their errors will propagate to the final estimation. However, this issue is addressed by the following optimality result.

\subsubsection{Theoretical guarantees for $\Psi_{dr}$}

We can show, under weak assumption on the quality of $\theta$, that $\Psi_{dr}$ has strong theoretical guarantees.
Let $o_p$ denote convergence in probability, define the $L_2(P)$ as $\|f\|_{L_2(P)} = (\int |f|^2 dP)^{1/2}$, where $P$ is the true distribution. We make the following assumption.
\begin{assumption}\label{assumption:4th-root-n}
Assume that both $P(Y|X,\theta)$ and $P(A=1|Y)$ converge at fourth-root-n rate i.e.
% \vspace*{-.1in}
\begin{equation}
\begin{aligned}
&\|P(Y|X,\theta) - P(Y|X)\|_{L_2(P)} = o_p(N^{-1/4})\\
&\|P(A=1|Y,\theta) - P(A=1|Y)\|_2 = o_p(N^{-1/4})
\end{aligned}
\end{equation} 
\end{assumption}
\noindent
\textbf{Justification:}
These assumptions (fourth-root-n rate of convergence) have been proven for neural networks \cite{riesznet}, which are consistent because of the universal approximation theorem, but tend to be biased because of regularization \cite{dml}.

We have the following optimality result:
\begin{theorem}  Under the assumption~\cref{assumption:4th-root-n} the DR estimator $\Psi_{dr}$ is asymptotically normal with 0-mean and the efficient influence function's variance:
% \vspace*{-.1in}
\begin{equation}
\sqrt{N}(\Psi_{dr}(\theta)(c) - P(Y=c)) \rightsquigarrow \mathcal{N}(0, \E[\phi(O)(c)^2])
\end{equation}
\label{theorem:dr}
\end{theorem}
The proof of theorem \cref{theorem:dr} is deferred to the supplemental material. This theorem states that $\Psi_{dr}$ is the most efficient regular estimator. 
Informally speaking, regularity means any other estimator that performs better than $\Psi_{dr}$ at one point, must do considerably worse at nearby points.%
% \footnote{Formally, an estimator $\Psi(\theta)$ of $\Psi(\theta_0)$ where $\theta_0$ is the truth is said to be regular if for every $h$
% \begin{equation}
% \sqrt{N} \left( \Psi(\theta) - \Psi\left( \theta_0 + \frac{h}{\sqrt{N}} \right) \right) \stackrel{\theta_0 + h / \sqrt{N}}{\rightsquigarrow} L_{\theta_0}
% \end{equation}
% where the convergence is in distribution under the law of  $\theta_0 + \frac{h}{\sqrt{N}}$, and $L$ is some asymptotic distribution that only depends on $\theta_0$ \cite{asymptoticstatistics}}

% As the efficient influence function $\phi$ has the smallest variance out of all regular and asymptotically linear (RAL) estimators of $P(Y)$, $\Psi_{dr}(\theta)$ is the most efficient RAL estimator. 
% }
To put this theorem into perspective, the sample mean $\frac{1}{n}\sum_i z_i$ is the most efficient estimator of the mean of a random variable $Z$, of which $z_i$s are unbiased samples. The OR estimator $\Psi_{or}$, which looks like a sample mean of $P(Y|X,\theta)$, is however potentially biased as $\theta$ is the model's approximation of the truth using finite data, and this bias slows the convergence of $\Psi_{or}$ if it does not go away quickly enough, for example if the first equation in assumption~\ref{assumption:4th-root-n} holds \cite{dml}. The same thing can happen to $P(A=1|Y,\theta)$. Thus, \cref{theorem:dr} shows that we can get an estimation quality as if we were using unbiased samples to estimate the mean.

\section{Experimental Results}
\begin{table*}[t]
\centering
\caption{Total Variation Distance on CIFAR-10-LT ($N_l = 500$, $M_l = 4000$) with different class imbalance ratios $\gamma_l$ and $\gamma_u$ under five different unlabeled class distributions.}
\label{tab:cifar10-tv}
\resizebox{\textwidth}{!}{
\begin{tabular}{lccccccccccc}
\toprule
& & \multicolumn{2}{c}{consistent} & \multicolumn{2}{c}{uniform} & \multicolumn{2}{c}{reversed} & \multicolumn{2}{c}{middle} & \multicolumn{2}{c}{head-tail} \\
\cmidrule(lr){3-4} \cmidrule(lr){5-6} \cmidrule(lr){7-8} \cmidrule(lr){9-10} \cmidrule(lr){11-12}
& & $\gamma_l = 150$ & $\gamma_l = 100$ & $\gamma_l = 150$ & $\gamma_l = 100$ & $\gamma_l = 150$ & $\gamma_l = 100$ & $\gamma_l = 150$ & $\gamma_l = 100$ & $\gamma_l = 150$ & $\gamma_l = 100$ \\
Model & Estimator & $\gamma_u = 150$ & $\gamma_u = 100$ & $\gamma_u = 1$ & $\gamma_u = 1$ & $\gamma_u = 1/150$ & $\gamma_u = 1/100$ & $\gamma_u = 150$ & $\gamma_u = 100$ & $\gamma_u = 150$ & $\gamma_u = 100$ \\
\midrule
Supervised & MLLS & 0.269 ± 0.252 & 0.038 ± 0.006 & 0.251 ± 0.046 & 0.255 ± 0.060 & 0.429 ± 0.028 & 0.493 ± 0.050 & 0.333 ± 0.042 & 0.320 ± 0.009 & 0.457 ± 0.034 & 0.444 ± 0.043 \\
Supervised & RLLS & 0.043 ± 0.001 & 0.044 ± 0.010 & 0.348 ± 0.034 & 0.305 ± 0.068 & 0.769 ± 0.016 & 0.678 ± 0.028 & 0.430 ± 0.008 & 0.368 ± 0.013 & 0.539 ± 0.018 & 0.503 ± 0.020 \\
\midrule
MLE & IPW & 0.027 ± 0.001 & 0.027 ± 0.000 & 0.319 ± 0.072 & 0.243 ± 0.010 & 0.674 ± 0.020 & 0.646 ± 0.041 & 0.438 ± 0.020 & 0.454 ± 0.026 & 0.547 ± 0.049 & 0.491 ± 0.059 \\
MLE & OR & 0.045 ± 0.004 & 0.042 ± 0.000 & 0.215 ± 0.026 & 0.203 ± 0.032 & 0.433 ± 0.017 & 0.395 ± 0.033 & 0.193 ± 0.006 & 0.209 ± 0.037 & 0.307 ± 0.147 & 0.249 ± 0.130 \\
MLE & DR & 0.090 ± 0.002 & 0.079 ± 0.000 & 0.407 ± 0.027 & 0.360 ± 0.007 & 0.425 ± 0.007 & 0.421 ± 0.029 & 0.256 ± 0.001 & 0.286 ± 0.031 & 0.435 ± 0.136 & 0.362 ± 0.122 \\
\midrule
EM & IPW & 0.035 ± 0.002 & 0.040 ± 0.001 & 0.021 ± 0.001 & 0.029 ± 0.015 & 0.303 ± 0.187 & 0.091 ± 0.010 & 0.119 ± 0.011 & 0.105 ± 0.022 & 0.104 ± 0.026 & 0.104 ± 0.051 \\
EM & OR & 0.037 ± 0.003 & 0.042 ± 0.002 & 0.016 ± 0.001 & 0.024 ± 0.012 & 0.269 ± 0.183 & 0.090 ± 0.008 & 0.122 ± 0.012 & 0.103 ± 0.022 & 0.072 ± 0.012 & 0.073 ± 0.024 \\
EM & DR & 0.034 ± 0.004 & 0.037 ± 0.001 & 0.014 ± 0.001 & 0.027 ± 0.020 & 0.264 ± 0.191 & 0.092 ± 0.005 & 0.111 ± 0.019 & 0.097 ± 0.026 & 0.077 ± 0.016 & 0.073 ± 0.028 \\
\midrule
SimPro & IPW & 0.070 ± 0.011 & 0.058 ± 0.000 & 0.046 ± 0.001 & 0.049 ± 0.005 & 0.254 ± 0.074 & 0.223 ± 0.098 & 0.097 ± 0.025 & 0.067 ± 0.002 & 0.105 ± 0.066 & 0.110 ± 0.079 \\
SimPro & OR & 0.071 ± 0.012 & 0.058 ± 0.000 & 0.045 ± 0.001 & 0.049 ± 0.006 & 0.040 ± 0.003 & 0.059 ± 0.017 & 0.074 ± 0.006 & 0.075 ± 0.002 & 0.033 ± 0.003 & 0.033 ± 0.003 \\
SimPro & DR & 0.017 ± 0.004 & 0.026 ± 0.001 & 0.019 ± 0.002 & 0.018 ± 0.003 & 0.039 ± 0.003 & 0.058 ± 0.025 & 0.091 ± 0.007 & 0.031 ± 0.001 & 0.015 ± 0.003 & 0.019 ± 0.007 \\
\bottomrule
\end{tabular}
}
\end{table*}

\begin{table*}[t]
\centering
\caption{Total Variation Distance on CIFAR-100-LT ($N_l = 50$, $M_l = 400$) with different class imbalance ratios $\gamma_l$ and $\gamma_u$ under five different unlabeled class distributions.}
\label{tab:cifar100-tv}
\resizebox{\textwidth}{!}{
\begin{tabular}{lccccccccccc}
\toprule
& & \multicolumn{2}{c}{consistent} & \multicolumn{2}{c}{uniform} & \multicolumn{2}{c}{reversed} & \multicolumn{2}{c}{middle} & \multicolumn{2}{c}{head-tail} \\
\cmidrule(lr){3-4} \cmidrule(lr){5-6} \cmidrule(lr){7-8} \cmidrule(lr){9-10} \cmidrule(lr){11-12}
& & $\gamma_l = 20$ & $\gamma_l = 10$ & $\gamma_l = 20$ & $\gamma_l = 10$ & $\gamma_l = 20$ & $\gamma_l = 10$ & $\gamma_l = 20$ & $\gamma_l = 10$ & $\gamma_l = 20$ & $\gamma_l = 10$ \\
Model & Estimator & $\gamma_u = 20$ & $\gamma_u = 10$ & $\gamma_u = 1$ & $\gamma_u = 1$ & $\gamma_u = 1/20$ & $\gamma_u = 1/10$ & $\gamma_u = 20$ & $\gamma_u = 10$ & $\gamma_u = 20$ & $\gamma_u = 10$ \\
\midrule
Supervised & MLLS & 0.707 ± 0.016 & 0.313 ± 0.100 & 0.445 ± 0.172 & 0.309 ± 0.119 & 0.383 ± 0.075 & 0.397 ± 0.006 & 0.570 ± 0.001 & 0.373 ± 0.107 & 0.543 ± 0.009 & 0.231 ± 0.057 \\
Supervised & RLLS & 0.520 ± 0.007 & 0.133 ± 0.003 & 0.337 ± 0.125 & 0.253 ± 0.082 & 0.424 ± 0.060 & 0.463 ± 0.003 & 0.454 ± 0.021 & 0.306 ± 0.074 & 0.460 ± 0.028 & 0.241 ± 0.040 \\
\midrule
MLE & IPW & 0.075 ± 0.000 & 0.071 ± 0.001 & 0.229 ± 0.001 & 0.167 ± 0.002 & 0.565 ± 0.005 & 0.443 ± 0.007 & 0.415 ± 0.000 & 0.311 ± 0.005 & 0.343 ± 0.000 & 0.280 ± 0.001 \\
MLE & OR & 0.065 ± 0.002 & 0.061 ± 0.001 & 0.200 ± 0.007 & 0.143 ± 0.001 & 0.526 ± 0.011 & 0.399 ± 0.023 & 0.360 ± 0.003 & 0.256 ± 0.012 & 0.328 ± 0.003 & 0.266 ± 0.005 \\
MLE & DR & 0.149 ± 0.019 & 0.145 ± 0.010 & 0.243 ± 0.004 & 0.214 ± 0.019 & 0.568 ± 0.005 & 0.464 ± 0.014 & 0.403 ± 0.014 & 0.309 ± 0.012 & 0.365 ± 0.007 & 0.320 ± 0.004 \\
\midrule
EM & IPW & 0.097 ± 0.008 & 0.092 ± 0.004 & 0.239 ± 0.007 & 0.179 ± 0.003 & 0.478 ± 0.012 & 0.329 ± 0.020 & 0.262 ± 0.016 & 0.202 ± 0.003 & 0.312 ± 0.002 & 0.227 ± 0.001 \\
EM & OR & 0.121 ± 0.007 & 0.108 ± 0.005 & 0.261 ± 0.007 & 0.189 ± 0.004 & 0.489 ± 0.013 & 0.335 ± 0.020 & 0.274 ± 0.016 & 0.211 ± 0.004 & 0.336 ± 0.003 & 0.235 ± 0.001 \\
EM & DR & 0.125 ± 0.005 & 0.111 ± 0.004 & 0.269 ± 0.007 & 0.194 ± 0.005 & 0.497 ± 0.010 & 0.336 ± 0.024 & 0.281 ± 0.019 & 0.219 ± 0.008 & 0.336 ± 0.007 & 0.233 ± 0.004 \\
\midrule
SimPro & IPW & 0.125 ± 0.001 & 0.100 ± 0.005 & 0.166 ± 0.007 & 0.141 ± 0.009 & 0.353 ± 0.023 & 0.261 ± 0.008 & 0.202 ± 0.003 & 0.158 ± 0.005 & 0.277 ± 0.009 & 0.197 ± 0.003 \\
SimPro & OR & 0.133 ± 0.005 & 0.100 ± 0.004 & 0.160 ± 0.007 & 0.138 ± 0.010 & 0.322 ± 0.014 & 0.253 ± 0.008 & 0.202 ± 0.003 & 0.156 ± 0.005 & 0.269 ± 0.006 & 0.191 ± 0.004 \\
SimPro & DR & 0.122 ± 0.003 & 0.106 ± 0.006 & 0.188 ± 0.009 & 0.149 ± 0.006 & 0.343 ± 0.023 & 0.257 ± 0.007 & 0.219 ± 0.010 & 0.172 ± 0.002 & 0.279 ± 0.007 & 0.198 ± 0.004 \\
\bottomrule
\end{tabular}
}
\end{table*}
\begin{table*}[t]
\centering
\caption{Top-1 accuracy (\%) on CIFAR-10-LT ($N_l = 500$, $M_l = 4000$) with different class imbalance ratios $\gamma_l$ and $\gamma_u$ under five different unlabeled class distributions. In most settings, our two stage algorithm improves SimPro (9 / 10) and BOAT (8 / 10). We use {\green green} to indicate when our plug-in improves and {\red red} when it degrades the base model.}
\label{tab:cifar10-acc}
\resizebox{\textwidth}{!}{
\begin{tabular}{lcccccccccc}
\toprule

& \multicolumn{2}{c}{consistent} & \multicolumn{2}{c}{uniform} & \multicolumn{2}{c}{reversed} & \multicolumn{2}{c}{middle} & \multicolumn{2}{c}{head-tail} \\
\cmidrule(lr){2-3} \cmidrule(lr){4-5} \cmidrule(lr){6-7} \cmidrule(lr){8-9} \cmidrule(lr){10-11}

& $\gamma_l = 150$ & $\gamma_l = 100$ & $\gamma_l = 150$ & $\gamma_l = 100$ & $\gamma_l = 150$ & $\gamma_l = 100$ & $\gamma_l = 150$ & $\gamma_l = 100$ & $\gamma_l = 150$ & $\gamma_l = 100$ \\
& $\gamma_u = 150$ & $\gamma_u = 100$ & $\gamma_u = 1$ & $\gamma_u = 1$ & $\gamma_u = 1/150$ & $\gamma_u = 1/100$ & $\gamma_u = 150$ & $\gamma_u = 100$ & $\gamma_u = 150$ & $\gamma_u = 100$ \\

\midrule

FixMatch & 62.9 $\pm$ 0.36 & 67.8 $\pm$ 1.13 & 67.6 $\pm$ 2.56 & 73.0 $\pm$ 3.81 & 59.9 $\pm$ 0.82 & 62.5 $\pm$ 0.94 & 64.3 $\pm$ 0.63 & 71.7 $\pm$ 0.46 & 58.3 $\pm$ 1.46 & 66.6 $\pm$ 0.87 \\
CReST+ & 67.5 $\pm$ 0.45 & 76.3 $\pm$ 0.86 & 74.9 $\pm$ 0.90 & 82.2 $\pm$ 1.53 & 62.0 $\pm$ 1.18 & 62.9 $\pm$ 1.39 & 58.5 $\pm$ 0.68 & 71.4 $\pm$ 0.60 & 59.3 $\pm$ 0.72 & 67.2 $\pm$ 0.48 \\
DASO & 70.1 $\pm$ 1.81 & 76.0 $\pm$ 0.37 & 83.1 $\pm$ 0.47 & 86.6 $\pm$ 0.84 & 64.0 $\pm$ 0.11 & 71.0 $\pm$ 0.95 & 69.0 $\pm$ 0.31 & 73.1 $\pm$ 0.68 & 70.5 $\pm$ 0.59 & 71.1 $\pm$ 0.32 \\
% w/ ACR$\dagger$ (Wei \& Gan, 2023) & 70.9 $\pm$ 0.37 & 76.1 $\pm$ 0.42 & 91.9 $\pm$ 0.02 & 92.5 $\pm$ 0.19 & 83.2 $\pm$ 0.39 & 85.2 $\pm$ 0.12 & 77.6 $\pm$ 0.20 & 79.3 $\pm$ 0.30 & 73.8 $\pm$ 0.83 & 79.3 $\pm$ 0.48 \\
% w/ SimPro & 74.2 $\pm$ 0.90 & 80.7 $\pm$ 0.30 & 93.6 $\pm$ 0.08 & 93.8 $\pm$ 0.10 & 83.5 $\pm$ 0.95 & 85.8 $\pm$ 0.48 & 82.6 $\pm$ 0.38 & 84.8 $\pm$ 0.54 & 81.0 $\pm$ 0.27 & 83.0 $\pm$ 0.36 \\
Supervised & 63.2 $\pm$ 0.14 & 66.0 $\pm$ 0.27 & 63.3 $\pm$ 0.28 & 65.8 $\pm$ 0.19 & 63.1 $\pm$ 0.19 & 65.9 $\pm$ 0.51 & 63.5 $\pm$ 0.22 & 65.8 $\pm$ 0.03 & 63.0 $\pm$ 0.18 & 66.4 $\pm$ 0.07 \\
\midrule
EM & 69.1 $\pm$ 1.29 & 73.8 $\pm$ 0.71 & 94.0 $\pm$ 0.08 & 93.2 $\pm$ 0.94 & 76.6 $\pm$ 2.72 & 82.2 $\pm$ 0.24 & 79.5 $\pm$ 0.35 & 81.6 $\pm$ 0.58 & 79.2 $\pm$ 0.50 & 79.8 $\pm$ 0.17 \\
\midrule
SimPro & 74.4 $\pm$ 0.71 & 79.7 $\pm$ 0.45 & 93.3 $\pm$ 0.10 & 93.3 $\pm$ 0.47 & 83.8 $\pm$ 0.80 & 84.1 $\pm$ 0.24 & 78.7 $\pm$ 0.30 & 84.2 $\pm$ 0.26 & 81.2 $\pm$ 0.20 & 82.0 $\pm$ 1.07 \\
% \midrule
SimPro+ & \green 77.8 $\pm$ 1.50 & \green 81.2 $\pm$ 0.39 & \green 93.7 $\pm$ 0.07 & \green 93.7 $\pm$ 0.24 & \red 83.3 $\pm$ 0.38 & \green 84.7 $\pm$ 0.78 & \green 79.2 $\pm$ 0.70 & \green 85.4 $\pm$ 0.66 & \green 81.3 $\pm$ 0.27 & \green 82.5 $\pm$ 0.56 \\
\midrule
BOAT & 80.5 $\pm$ 0.39 & 83.3 $\pm$ 0.27 & 93.9 $\pm$ 0.03 & 94.1 $\pm$ 0.10 & 79.7 $\pm$ 0.25 & 81.1 $\pm$ 0.15 & 79.7 $\pm$ 1.15 & 81.6 $\pm$ 0.09 & 79.4 $\pm$ 0.44 & 80.9 $\pm$ 0.16 \\
% \midrule
BOAT+ & \green 81.6 $\pm$ 0.15 & \green 83.8 $\pm$ 0.04 & \red 93.7 $\pm$ 0.23 & 94.1 $\pm$ 0.17 & \green 80.4 $\pm$ 0.71 & \green 81.7 $\pm$ 0.38 & \green 80.3 $\pm$ 0.28 & \green 83.1 $\pm$ 0.45 & \green 79.7 $\pm$ 0.29 & \green 81.0 $\pm$ 0.36 \\
\bottomrule
\end{tabular}
}
\end{table*}

\begin{table*}[t]
\centering
\caption{Top-1 accuracy (\%) on CIFAR-100-LT ($N_l = 50$, $M_l = 400$) with different class imbalance ratios $\gamma_l$ and $\gamma_u$ under five different unlabeled class distributions. Despite poor estimation in stage 1, our approach does not degrade the accuracy for most of the settings. We use {\green green} to indicate when our plug-in improves and {\red red} when it degrades the base method.}
\label{tab:cifar100-acc}
\resizebox{\textwidth}{!}{
\begin{tabular}{lccccccccccc}
\toprule

& \multicolumn{2}{c}{consistent} & \multicolumn{2}{c}{uniform} & \multicolumn{2}{c}{reversed} & \multicolumn{2}{c}{middle} & \multicolumn{2}{c}{head-tail} \\
\cmidrule(lr){2-3} \cmidrule(lr){4-5} \cmidrule(lr){6-7} \cmidrule(lr){8-9} \cmidrule(lr){10-11}

& $\gamma_l = 20$ & $\gamma_l = 10$ & $\gamma_l = 20$ & $\gamma_l = 10$ & $\gamma_l = 20$ & $\gamma_l = 10$ & $\gamma_l = 20$ & $\gamma_l = 10$ & $\gamma_l = 20$ & $\gamma_l = 10$ \\
& $\gamma_u = 20$ & $\gamma_u = 10$ & $\gamma_u = 1$ & $\gamma_u = 1$ & $\gamma_u = 1/20$ & $\gamma_u = 1/10$ & $\gamma_u = 20$ & $\gamma_u = 10$ & $\gamma_u = 20$ & $\gamma_u = 10$ \\

\midrule
% FixMatch & 40.0 $\pm$ 0.96 & 45.2 $\pm$ 0.55 & 39.6 $\pm$ 1.16 & \\
% CReST+ & 40.1 $\pm$ 1.28 & 44.5 $\pm$ 0.94 & 37.6 $\pm$ 0.88 & \\
% DASO & 43.0 $\pm$ 0.15 & 49.8 $\pm$ 0.24 & 49.4 $\pm$ 0.93 & \\
Supervised & 32.4 $\pm$ 0.40 & 38.4 $\pm$ 0.18 & 32.7 $\pm$ 0.25 & 38.0 $\pm$ 0.22 & 32.5 $\pm$ 0.51 & 38.4 $\pm$ 0.43 & 32.3 $\pm$ 0.08 & 37.9 $\pm$ 0.43 & 32.1 $\pm$ 0.33 & 38.2 $\pm$ 0.38 \\
% \midrule
EM & 42.4 $\pm$ 0.43 & 49.6 $\pm$ 0.30 & 50.9 $\pm$ 0.27 & 58.0 $\pm$ 0.35 & 42.1 $\pm$ 0.16 & 49.8 $\pm$ 0.47 & 42.8 $\pm$ 0.41 & 49.6 $\pm$ 0.36 & 41.5 $\pm$ 1.26 & 49.5 $\pm$ 0.18 \\
\midrule
SimPro & 42.5 $\pm$ 0.58 & 49.6 $\pm$ 0.22 & 51.7 $\pm$ 0.22 & 58.1 $\pm$ 0.53 & 44.9 $\pm$ 0.21 & 51.8 $\pm$ 0.42 & 42.7 $\pm$ 0.06 & 49.8 $\pm$ 0.45 & 43.3 $\pm$ 0.76 & 50.9 $\pm$ 0.19 \\
% \midrule
SimPro+ & \green 42.8 $\pm$ 0.49 & \green 50.1 $\pm$ 0.33 & \red 51.6 $\pm$ 0.63 & \red 57.8 $\pm$ 0.38 & \red 44.7 $\pm$ 0.51 & \red 51.4 $\pm$ 0.88 & \green 43.4 $\pm$ 0.58 & \green 50.4 $\pm$ 0.28 & \green 43.8 $\pm$ 0.50 & \red 50.7 $\pm$ 0.76 \\
\midrule
BOAT & 43.7 $\pm$ 0.16 & 51.4 $\pm$ 0.32 & 55.1 $\pm$ 0.95 & 60.5 $\pm$ 0.15 & 43.1 $\pm$ 0.49 & 52.7 $\pm$ 0.23 & 43.6 $\pm$ 0.19 & 51.4 $\pm$ 0.39 & 43.9 $\pm$ 0.42 & 51.4 $\pm$ 0.14 \\
% \midrule
BOAT+ & \green 44.8 $\pm$ 0.13 & 51.4 $\pm$ 0.51 & \red 53.8 $\pm$ 0.32 & 60.5 $\pm$ 0.69 & \green 43.4 $\pm$ 0.56 & \red 52.4 $\pm$ 0.36 & \green 43.9 $\pm$ 0.59 & \red 50.8 $\pm$ 0.09 & \red 43.6 $\pm$ 0.50 & \green 51.9 $\pm$ 0.49 \\
\bottomrule
\end{tabular}
}
\end{table*}

We perform experiments for each stage of our algorithm. In the first stage, we compare among various methods to estimate the unlabeled class distribution $P(Y|A=0)$, showing that SimPro + DR performs well. In the second stage, we freeze the unlabeled class distribution, using our best estimator  SimPro + DR, and plug it into 2 SOTA semi-supervised learning algorithms, SimPro and BOAT~\cite{boat}. We show that this simple procedure improves the existing methods, and is even capable of improving the original SimPro when used for both stages.

% \textbf{Datasets} We adopt 4 standard benchmarks used frequently in other semi-supervised learning work: CIFAR-10, CIFAR-100~\cite{cifar}, STL-10~\cite{stl10} and Imagenet-127~\cite{cossl}. To match our RTSSL setting, we create long-tailed labeled and unlabeled sets from CIFAR-10 and CIFAR-100. Specifically, we use $\gamma_l$ and $n_1$ to denote the imbalance ratio and the head class's number of samples of the labeled data, the remaining class's size is computed as $n_c = n_1 \times \gamma_l^{-\frac{c-1}{C-1}}$ and likewise, $\gamma_u$ and $m_1$ of the unlabeled data. For CIFAR-10, we fix $n_1=500$ and $m_1=4000$. We test 2 different configurations $\gamma_l=\gamma_c=150$ and $\gamma_l=\gamma_c=100$. We further permute classes the unlabeled sets in 5 ways: consistent, uniform, reversed, middle and headtail, similar to \cite{simpro} and visualized in figure~\ref{fig:distribution}, which results in 10 different datasets in total. Similarly for CIFAR-100, we fix $n_1=500$ and $m_1=4000$, use 2 configurations $\gamma_l=\gamma_c=20$ and $\gamma_l=\gamma_c=10$, and permute the classes in 5 ways, resulting in 10 datasets as well. For STL-10, the unlabeled set has no ground truth labels, therefore we use all samples in the head class and set the imbalance ratio $\gamma_l$ to $10$ or $20$. Imagenet-127 is a naturally long-tailed dataset with 127 classes. We train on 32x32 and 64x64 image resolutions following ~\cite{cossl}.

\textbf{Datasets} We evaluate our method on four standard semi-supervised learning benchmarks: CIFAR-10, CIFAR-100~\cite{cifar}, STL-10~\cite{stl10}, and Imagenet-127~\cite{cossl}. To simulate RTSSL, we construct long-tailed labeled and unlabeled sets for CIFAR-10 and CIFAR-100. The labeled data follows an imbalance ratio $\gamma_l$ with head class size $n_1$, while the remaining class sizes are computed as $n_c = n_1 \times \gamma_l^{-\frac{c-1}{C-1}}$. The unlabeled data follows a similar setup with $\gamma_u$ and $m_1$.  

For CIFAR-10, we set $n_1 = 500$, $m_1 = 4000$, and test two configurations: $\gamma_l = \gamma_u = 150$ and $\gamma_l = \gamma_u = 100$. We generate 10 datasets by permuting the unlabeled class distributions in five ways: \textit{consistent, uniform, reversed, middle}, and \textit{head-tail}, as in~\cite{simpro}. CIFAR-100 follows the same setup with $n_1 = 50$, $m_1 = 400$, and $\gamma_l, \gamma_u$ values of 20 and 10.  

For STL-10, where unlabeled data lacks ground-truth labels, we use all head-class samples and set $\gamma_l$ to 10 or 20. Imagenet-127 is naturally long-tailed with 127 classes, and we train on 32$\times$32 and 64$\times$64 resolutions as in~\cite{cossl}.

\paragraph{Training.} We follow the implementation and hyperparameter settings of \cite{simpro}. We defer these details in \cref{subsec:training-setting}. One important exception is that for Imagenet-127, we use the smaller Wide ResNet-28-2 in stage 1 and the larger ResNet-50 for stage 2, to demonstrate that a smaller model is sufficient for distribution estimation.

\begin{table}[t]
\small
\centering
\caption{Top-1 Accuracy (\%) on STL-10. Our two-stage algorithms improves both SimPro and BOAT for both settings.}
\label{tab:stl10-acc}
% \resizebox{\linewidth}{!}{
\begin{tabular}{lcc}
\toprule
Method & $\gamma_l=10$ & $\gamma_l=20$ \\ \hline
Supervised & 73.9 $\pm$ 0.57 & 70.4 $\pm$ 0.95 \\
\midrule
MLE & 67.6 $\pm$ 0.57 & 58.9 $\pm$ 4.05 \\
\midrule
EM & 84.9 $\pm$ 0.14 & 83.6 $\pm$ 0.25 \\
\midrule
SimPro & 82.4 $\pm$ 1.57 & 80.5 $\pm$ 0.96 \\
SimPro+ & \green 83.9 $\pm$ 0.76 & \green 82.7 $\pm$ 0.86 \\
\midrule
BOAT & 83.8 $\pm$ 0.20 & 82.0 $\pm$ 0.34 \\
BOAT+ & \green 84.1 $\pm$ 0.38 & \green 82.4 $\pm$ 0.10 \\
\bottomrule
\end{tabular}
\end{table}

\begin{table}[t]
\small
\centering
\caption{Top-1 Accuracy (\%) on Imagenet-127. Our two-stage approach improves both SimPro and BOAT for both resolutions.}
\label{tab:imagenet-127-acc}
% \resizebox{\linewidth}{!}{
\begin{tabular}{lcc}
\toprule
Method & $32 \times 32$ & $64 \times 64$ \\ \hline
SimPro & 54.8 & 63.7 \\
SimPro+ & \green 55.1 & \green 64.2 \\
\midrule
BOAT & 51.6 & 58.7 \\
BOAT+ & \green 52.0 & \green 59.2 \\

\bottomrule
\end{tabular}
% }
\end{table}

\begin{table}[t]
\small\centering
\caption{Total Variation Distance on Imagenet-127}
\label{tab:imagenet-127-tv}
% \resizebox{\linewidth}{!}{
\begin{tabular}{cccc}
\toprule
Method & Estimator & $32 \times 32$ & $64 \times 64$ \\ \hline
MLE & IPW  & 0.103 $\pm$ 0.034 & 0.051 $\pm$ 0.000 \\
MLE & OR  & 0.153 $\pm$ 0.052 & 0.041 $\pm$ 0.000 \\
MLE & DR  & \green 0.100 $\pm$ 0.029 & \green 0.075 $\pm$ 0.003 \\
\midrule
EM & IPW  & 0.141 $\pm$ 0.006 & 0.163 $\pm$ 0.010 \\
EM & OR  & 0.205 $\pm$ 0.006 & 0.236 $\pm$ 0.011 \\
EM & DR  & \green 0.024 $\pm$ 0.001 & \green 0.042 $\pm$ 0.004 \\
\midrule
SimPro & IPW  & 0.041 $\pm$ 0.012 & 0.224 $\pm$ 0.040 \\
SimPro & OR  & 0.036 $\pm$ 0.014 & 0.291 $\pm$ 0.079 \\
SimPro & DR  & \green 0.017 $\pm$ 0.000 & \green 0.037 $\pm$ 0.004 \\
\bottomrule
\end{tabular}
% }
\end{table}

\subsection{Better results on label distribution} 
\label{subsec:label}
We have mentioned various ways throughout the papers to estimate the unlabeled class distribution. In what follows, method consists of a model, which is how the learning is done, and an estimator, which is how the final distribution is estimated using parameters learned from the model.

%\begin{enumerate}
%\item 
\noindent
\textbf{Supervised}. The model is trained on the labeled set only and used to estimate the unlabeled class distribution \cite{unifiedlabelshift}. 2 successful estimators for this setting are \textbf{RLLS} \cite{rlls} and \textbf{MLLS} \cite{mlls}. 

%\item 
\noindent\textbf{MLE}. The model is trained by directly maximizing the likelihood \cref{eq:likelihood}. We also use the decomposition $P(Y|X)$ and $P(A|Y)$, and write the unlabeled term as $P(A=0, X) = \sum_{c} P(Y=c|X) P(A=0|Y=c)$, which enables gradient descent training on these parameters. This is also the MLE method to estimate $P(A|Y)$ in \cite{arelabelsinformative}.

%\item 
\noindent\textbf{EM}. We further test the EM algorithm in \cref{subsec:em}. In particular we also use strong and weak augmentations similar to FixMatch, but not the pseudo labeling operator. Confidence thresholding removes the soft predictions of the non-dominant classes, which may be better to keep since our target of the first stage is the global class statistics. We also try 3 estimators on this model.

%\item 
\noindent\textbf{SimPro} \cite{simpro} can be seen as our previous EM but also with FixMatch's confidence thresholding and logit adjustment loss in \cref{subsec:simpro}. Confidence thresholding is a powerful regularization technique that encodes the assumption that classes are well separated \cite{entropyminimization}, but can introduce bias to the estimation, which justifies the use of DR.
%\end{enumerate}

% For semi-supervised methods MLE, EM and SimPro, as we now have additional information on the missingness mechanism, we can use 3 estimators OR, IPW and DR presented in \cref{subsec:2-stage}

Results on \cref{tab:cifar10-tv} presents the performance of various models and estimators on CIFAR-10. We can see that SimPro + DR performs best. In contrast, SimPro + OR, SimPro's original way of estimating $P(Y|A=0)$, and SimPro + IPW tend to underperform EM on the consistent and uniform datasets. The consistent setting is worth noting, since it arises when data is sampled uniformly at random for labeling,  representative of a large number of real world situations. EM is competitive to SimPro as well even without pseudo labeling, but overall we found this regularization to offer significant gains in the reversed, middle and head-tail settings. Finally, Supervised with either MLLS or RLLS estimators performs much worse than the semi-supervise methods.

\cref{tab:imagenet-127-tv} aligns with the observations  made in \cref{tab:cifar10-tv}. In particular, SimPro + DR is the best method for class distribution estimation of the much larger Imagenet-127. We also found that a small neural network and a small image resolution is sufficient for the distribution estimation of the much larger dataset Imagenet-127. This matches our intuition that the finite-dimensional parameter is easier to learn.

\cref{tab:cifar100-tv} shows that most methods understandably struggle to estimate the class distributions in CIFAR-100. This is because there are few samples in each class, the head class has 10 times less samples while the number of classes multiplies 10 times compared to CIFAR-10. We see here that SimPro + DR is not the best method, but the relative gap between estimators are small.

% Among the models, the supervised baseline do not perform well even in the consistent setting, showing that when unlabeled data is available during training, learning from them can be valuable for class distribution estimation, especially in the cases with little labeled data like ours. Both the MLE and supervised models perform badly on the reversed, middle and head-tail settings

% Among the estimators, we see that DR boosts the performance of SimPro and EM in CIFAR-10, and of all semi-supervised models in Imagenet-127. It does not improve MLE on CIFAR-10, and it does not improve on CIFAR-100. However, for most of the time, the decrease is not much. In constrast, IPW estimators can be significantly worse, for example in the reversed setting of CIFAR-10, where the distance is $0.254$ for $\gamma_l=150$ and $0.233$ for $\gamma_l=100$, compared to OR's 0.040 and 0.059. 

% Both the MLE and supervised models perform badly on the reversed, middle and head-tail settings. EM does a decent job, though not as well as SimPro, on all 5 distribution settings of CIFAR-10. However, on Imagenet-127, EM without DR performs worse than MLE.

% We note that the performance on DR is similar to OR in these cases, showing that DR has a double robustness property. While IPW only relies on the finite-dimensional $P(A|Y)$, which intuitively is easy to estimate, we found that the inverse probability weight can nevertheless be unstable when some probabilities are small, and this is where DR shows its strength by combining both IPW and OR.

\subsection{Two-stage algorithm improves accuracy}

In the second stage of our algorithm, we freeze our estimation and plug it in SimPro and BOAT. We denote SimPro+ and BOAT+ for algorithms that use our first stage estimate.

\cref{tab:cifar10-acc} shows that for CIFAR-10 SimPro+ and BOAT+ improve over their original versions across most settings, leading to large improvements in both the consistent and middle class distribution settings. In particular, our two-stage approach improves SimPro in 9 / 10 settings and BOAT in 8 / 10 settings.
We also observe consistent improvements ove both base algorithms, SimPro and BOAT, for several other datasets. \cref{tab:stl10-acc} demonstrates improvements for 2 / 2 class imbalance ratios in STL-10 and \cref{tab:imagenet-127-acc} for 2 / 2  different resolutions of ImageNet-127.

We also evaluate on CIFAR-100 for multiple unlabeled  class distribution settings and with mediocre class label distribution estimates in stage 1, demonstrate no degradation in accuracy in stage 2. As shown in \cref{tab:cifar100-acc}, the two stage algorithm with a mediocre stage 1 estimation leads to parity with the baseline. Stage 2 provides small improvements in 5 / 10 settings for SimPro and in 4 / 10 (with 2 ties) for BOAT.

\subsection{Ablation Study: Alternative implementations.}
\label{subsec:ablation-1}
In this section, we ablate on our 2-stage choice. Specifically, we consider 2 alternative implementations:
\paragraph{\textbf{Doubly-robust risk}}  
This approach is \cite{arelabelsinformative, onnonrandommissinglabels}, as discussed in \cref{sec:background}. we consider the doubly-robust risk as our training loss. We use the missingness mechanism estimation from stage-1 of SimPro+ for fair comparison. \cref{eq:dr-risk} is used for training in which the pseudo-labeling operators can be applied straightforwardly. More detail in \cref{subsec:dr-risk}
\paragraph{\textbf{Batch-update doubly-robust $P(Y|A)$}} Different from SimPro+, here we remove the first stage and instead update our doubly robust estimation of the unlabeled class distribution using a moving average of the batch statistics.

\cref{tab:cifar10-ablation-1} shows that the batch-update version of SimPro+ is significantly worse on the consistent and uniform settings, while the doubly-robust risk is worst overall, especially in the reversed setting where $P(A|Y)$ is very small for the labeled tail classes, causing instability issues during training. In conclusion, our 2-stage approach is the best.

\begin{table}[t]
\small
\centering
\caption{Top-1 Accuracy (\%) on CIFAR-10. We compare our 2-stage SimPro+ with 1) an 1-stage alternative that updates and uses the doubly-robust estimation on-the-fly and 2) SimPro with doubly-robust risk. We use $\gamma_l=150$. {\green green} color indicates that our method performs best.}
\label{tab:cifar10-ablation-1}
\resizebox{\linewidth}{!}{
\begin{tabular}{lccccc}
\toprule
Method & consistent & uniform & reversed & middle & headtail\\ \hline
SimPro+ & \green 77.8 & \green 93.7 & \green 83.3 & \green 79.2 & \green 81.3 \\
batch-update & 71.9 & 91.4 & 82.6 & 78.6 & 81.2 \\
DR-risk & 72.1 & 89.8 & 67.1 & 75.6 & 79.5 \\
\bottomrule
\end{tabular}
}
\end{table}

\section{Conclusion}
we addressed the challenge of limited information about the class distribution in unlabeled data for semi-supervised learning. We propose to explicitly estimate the unlabeled class distribution as an initial step then pass it to existing pseudo labeling approaches.
Leveraging connecting to non-ignorable missingness, we introduce the doubly robust estimator which has strong theoretical guarantee for the distribution estimation. We show improved results on 3 different datasets, CIFAR-10, STL-10, and Imagenet-127, and that even inaccurate class label distributions do not lead to degraded accuracy in CIFAR-10.

\newpage
\section*{Impact Statement}

This paper presents work whose goal is to advance the field of 
Machine Learning. There are many potential societal consequences 
of our work, none which we feel must be specifically highlighted here.

\bibliography{main}

\begin{thebibliography}{55}
\providecommand{\natexlab}[1]{#1}
\providecommand{\url}[1]{\texttt{#1}}
\expandafter\ifx\csname urlstyle\endcsname\relax
  \providecommand{\doi}[1]{doi: #1}\else
  \providecommand{\doi}{doi: \begingroup \urlstyle{rm}\Url}\fi

\bibitem[Alexandari et~al.(2020)Alexandari, Kundaje, and Shrikumar]{mllsishardtobeat}
Alexandari, A., Kundaje, A., and Shrikumar, A.
\newblock Maximum likelihood with bias-corrected calibration is hard-to-beat at label shift adaptation.
\newblock In \emph{International Conference on Machine Learning}, pp.\  222--232. PMLR, 2020.

\bibitem[Arazo et~al.(2020)Arazo, Ortego, Albert, O’Connor, and McGuinness]{confirmationbias}
Arazo, E., Ortego, D., Albert, P., O’Connor, N.~E., and McGuinness, K.
\newblock Pseudo-labeling and confirmation bias in deep semi-supervised learning.
\newblock In \emph{2020 International joint conference on neural networks (IJCNN)}, pp.\  1--8. IEEE, 2020.

\bibitem[Azizzadenesheli et~al.(2019)Azizzadenesheli, Liu, Yang, and Anandkumar]{rlls}
Azizzadenesheli, K., Liu, A., Yang, F., and Anandkumar, A.
\newblock Regularized learning for domain adaptation under label shifts.
\newblock \emph{arXiv preprint arXiv:1903.09734}, 2019.

\bibitem[Berthelot et~al.(2019{\natexlab{a}})Berthelot, Carlini, Cubuk, Kurakin, Sohn, Zhang, and Raffel]{remixmatch}
Berthelot, D., Carlini, N., Cubuk, E.~D., Kurakin, A., Sohn, K., Zhang, H., and Raffel, C.
\newblock Remixmatch: Semi-supervised learning with distribution alignment and augmentation anchoring.
\newblock \emph{arXiv preprint arXiv:1911.09785}, 2019{\natexlab{a}}.

\bibitem[Berthelot et~al.(2019{\natexlab{b}})Berthelot, Carlini, Goodfellow, Papernot, Oliver, and Raffel]{mixmatch}
Berthelot, D., Carlini, N., Goodfellow, I., Papernot, N., Oliver, A., and Raffel, C.~A.
\newblock Mixmatch: A holistic approach to semi-supervised learning.
\newblock \emph{Advances in neural information processing systems}, 32, 2019{\natexlab{b}}.

\bibitem[Buda et~al.(2018)Buda, Maki, and Mazurowski]{longtailedlearning}
Buda, M., Maki, A., and Mazurowski, M.~A.
\newblock A systematic study of the class imbalance problem in convolutional neural networks.
\newblock \emph{Neural networks}, 106:\penalty0 249--259, 2018.

\bibitem[Chapelle et~al.(2009)Chapelle, Scholkopf, and Zien]{chapelle2009semi}
Chapelle, O., Scholkopf, B., and Zien, A.
\newblock Semi-supervised learning (chapelle, o. et al., eds.; 2006)[book reviews].
\newblock \emph{IEEE Transactions on Neural Networks}, 20\penalty0 (3):\penalty0 542--542, 2009.

\bibitem[Chernozhukov et~al.(2018)Chernozhukov, Chetverikov, Demirer, Duflo, Hansen, Newey, and Robins]{dml}
Chernozhukov, V., Chetverikov, D., Demirer, M., Duflo, E., Hansen, C., Newey, W., and Robins, J.
\newblock Double/debiased machine learning for treatment and structural parameters, 2018.

\bibitem[Chernozhukov et~al.(2022)Chernozhukov, Newey, Quintas-Mart{\i}nez, and Syrgkanis]{riesznet}
Chernozhukov, V., Newey, W., Quintas-Mart{\i}nez, V.~M., and Syrgkanis, V.
\newblock Riesznet and forestriesz: Automatic debiased machine learning with neural nets and random forests.
\newblock In \emph{International Conference on Machine Learning}, pp.\  3901--3914. PMLR, 2022.

\bibitem[Coates et~al.(2011)Coates, Ng, and Lee]{stl10}
Coates, A., Ng, A., and Lee, H.
\newblock An analysis of single-layer networks in unsupervised feature learning.
\newblock In \emph{Proceedings of the fourteenth international conference on artificial intelligence and statistics}, pp.\  215--223. JMLR Workshop and Conference Proceedings, 2011.

\bibitem[Cui et~al.(2019)Cui, Jia, Lin, Song, and Belongie]{cui-effective}
Cui, Y., Jia, M., Lin, T.-Y., Song, Y., and Belongie, S.
\newblock Class-balanced loss based on effective number of samples.
\newblock In \emph{Proceedings of the IEEE/CVF conference on computer vision and pattern recognition}, pp.\  9268--9277, 2019.

\bibitem[Du et~al.(2024)Du, Han, and Huang]{simpro}
Du, C., Han, Y., and Huang, G.
\newblock Simpro: A simple probabilistic framework towards realistic long-tailed semi-supervised learning.
\newblock \emph{arXiv preprint arXiv:2402.13505}, 2024.

\bibitem[Duan et~al.(2022)Duan, Qi, Wang, Zhou, and Shi]{rda}
Duan, Y., Qi, L., Wang, L., Zhou, L., and Shi, Y.
\newblock Rda: Reciprocal distribution alignment for robust semi-supervised learning.
\newblock In \emph{European Conference on Computer Vision}, pp.\  533--549. Springer, 2022.

\bibitem[Duan et~al.(2023)Duan, Zhao, Qi, Zhou, Wang, and Shi]{prg4ssl}
Duan, Y., Zhao, Z., Qi, L., Zhou, L., Wang, L., and Shi, Y.
\newblock Towards semi-supervised learning with non-random missing labels.
\newblock In \emph{Proceedings of the IEEE/CVF International Conference on Computer Vision}, pp.\  16121--16131, 2023.

\bibitem[Fan et~al.(2022)Fan, Dai, Kukleva, and Schiele]{cossl}
Fan, Y., Dai, D., Kukleva, A., and Schiele, B.
\newblock Cossl: Co-learning of representation and classifier for imbalanced semi-supervised learning.
\newblock In \emph{Proceedings of the IEEE/CVF conference on computer vision and pattern recognition}, pp.\  14574--14584, 2022.

\bibitem[Foster \& Syrgkanis(2023)Foster and Syrgkanis]{foster2023orthogonal}
Foster, D.~J. and Syrgkanis, V.
\newblock Orthogonal statistical learning.
\newblock \emph{The Annals of Statistics}, 51\penalty0 (3):\penalty0 879--908, 2023.

\bibitem[Gan et~al.(2024)Gan, Wei, and Zhang]{boat}
Gan, K., Wei, T., and Zhang, M.-L.
\newblock Boosting consistency in dual training for long-tailed semi-supervised learning.
\newblock \emph{arXiv preprint arXiv:2406.13187}, 2024.

\bibitem[Garg et~al.(2020)Garg, Wu, Balakrishnan, and Lipton]{unifiedlabelshift}
Garg, S., Wu, Y., Balakrishnan, S., and Lipton, Z.
\newblock A unified view of label shift estimation.
\newblock \emph{Advances in Neural Information Processing Systems}, 33:\penalty0 3290--3300, 2020.

\bibitem[Grandvalet \& Bengio(2004)Grandvalet and Bengio]{entropyminimization}
Grandvalet, Y. and Bengio, Y.
\newblock Semi-supervised learning by entropy minimization.
\newblock \emph{Advances in neural information processing systems}, 17, 2004.

\bibitem[Hu et~al.(2022)Hu, Niu, Miao, Hua, and Zhang]{onnonrandommissinglabels}
Hu, X., Niu, Y., Miao, C., Hua, X.-S., and Zhang, H.
\newblock On non-random missing labels in semi-supervised learning.
\newblock \emph{arXiv preprint arXiv:2206.14923}, 2022.

\bibitem[Ibrahim \& Lipsitz(1996)Ibrahim and Lipsitz]{ibrahim1996parameter}
Ibrahim, J.~G. and Lipsitz, S.~R.
\newblock Parameter estimation from incomplete data in binomial regression when the missing data mechanism is nonignorable.
\newblock \emph{Biometrics}, pp.\  1071--1078, 1996.

\bibitem[Kallus(2020)]{kallus2020deepmatch}
Kallus, N.
\newblock Deepmatch: Balancing deep covariate representations for causal inference using adversarial training.
\newblock In \emph{International Conference on Machine Learning}, pp.\  5067--5077. PMLR, 2020.

\bibitem[Kennedy(2023)]{kennedy2023towards}
Kennedy, E.~H.
\newblock Towards optimal doubly robust estimation of heterogeneous causal effects.
\newblock \emph{Electronic Journal of Statistics}, 17\penalty0 (2):\penalty0 3008--3049, 2023.

\bibitem[Kennedy(2024)]{kennedy-dr}
Kennedy, E.~H.
\newblock Semiparametric doubly robust targeted double machine learning: a review.
\newblock \emph{Handbook of Statistical Methods for Precision Medicine}, pp.\  207--236, 2024.

\bibitem[Kim et~al.(2020)Kim, Hur, Park, Yang, Hwang, and Shin]{kim2020distribution}
Kim, J., Hur, Y., Park, S., Yang, E., Hwang, S.~J., and Shin, J.
\newblock Distribution aligning refinery of pseudo-label for imbalanced semi-supervised learning.
\newblock \emph{Advances in neural information processing systems}, 33:\penalty0 14567--14579, 2020.

\bibitem[Krizhevsky \& Hinton(2009)Krizhevsky and Hinton]{cifar}
Krizhevsky, A. and Hinton, G.
\newblock Learning multiple layers of features from tiny images.
\newblock Technical Report~0, University of Toronto, Toronto, Ontario, 2009.
\newblock URL \url{https://www.cs.toronto.edu/~kriz/learning-features-2009-TR.pdf}.

\bibitem[Laine \& Aila(2016)Laine and Aila]{temporalensembling}
Laine, S. and Aila, T.
\newblock Temporal ensembling for semi-supervised learning.
\newblock \emph{arXiv preprint arXiv:1610.02242}, 2016.

\bibitem[Lee et~al.(2013)]{pseudolabeling}
Lee, D.-H. et~al.
\newblock Pseudo-label: The simple and efficient semi-supervised learning method for deep neural networks.
\newblock In \emph{Workshop on challenges in representation learning, ICML}, volume~3, pp.\  896. Atlanta, 2013.

\bibitem[Lee et~al.(2021)Lee, Shin, and Kim]{abc}
Lee, H., Shin, S., and Kim, H.
\newblock Abc: Auxiliary balanced classifier for class-imbalanced semi-supervised learning.
\newblock \emph{Advances in Neural Information Processing Systems}, 34:\penalty0 7082--7094, 2021.

\bibitem[Lipton et~al.(2018)Lipton, Wang, and Smola]{bbse}
Lipton, Z., Wang, Y.-X., and Smola, A.
\newblock Detecting and correcting for label shift with black box predictors.
\newblock In \emph{International conference on machine learning}, pp.\  3122--3130. PMLR, 2018.

\bibitem[Ma et~al.(2024)Ma, Elezi, Deng, Dong, and Xu]{cpe}
Ma, C., Elezi, I., Deng, J., Dong, W., and Xu, C.
\newblock Three heads are better than one: Complementary experts for long-tailed semi-supervised learning.
\newblock In \emph{Proceedings of the AAAI Conference on Artificial Intelligence}, volume~38, pp.\  14229--14237, 2024.

\bibitem[Menon et~al.(2020)Menon, Jayasumana, Rawat, Jain, Veit, and Kumar]{logitadjustment}
Menon, A.~K., Jayasumana, S., Rawat, A.~S., Jain, H., Veit, A., and Kumar, S.
\newblock Long-tail learning via logit adjustment.
\newblock \emph{arXiv preprint arXiv:2007.07314}, 2020.

\bibitem[Miller \& Futoma(2023)Miller and Futoma]{labelshift-nonignorable}
Miller, A.~C. and Futoma, J.
\newblock Label shift estimators for non-ignorable missing data.
\newblock \emph{arXiv preprint arXiv:2310.18261}, 2023.

\bibitem[Neal \& Hinton(1998)Neal and Hinton]{neal1998view}
Neal, R.~M. and Hinton, G.~E.
\newblock A view of the em algorithm that justifies incremental, sparse, and other variants.
\newblock In \emph{Learning in graphical models}, pp.\  355--368. Springer, 1998.

\bibitem[Nguyen et~al.(2024)Nguyen, Wang, Kim, and Sabuncu]{minh-gpa}
Nguyen, M., Wang, A.~Q., Kim, H., and Sabuncu, M.~R.
\newblock Adapting to shifting correlations with unlabeled data calibration.
\newblock \emph{arXiv preprint arXiv:2409.05996}, 2024.

\bibitem[Oh et~al.(2022)Oh, Kim, and Kweon]{oh2022daso}
Oh, Y., Kim, D.-J., and Kweon, I.~S.
\newblock {DASO}: Distribution-aware semantics-oriented pseudo-label for imbalanced semi-supervised learning.
\newblock In \emph{Proceedings of the IEEE/CVF conference on computer vision and pattern recognition}, pp.\  9786--9796, 2022.

\bibitem[Pham et~al.(2023)Pham, Hirshberg, Huynh-Pham, Santacatterina, Lim, and Zabih]{pham2023stable}
Pham, K., Hirshberg, D.~A., Huynh-Pham, P.-M., Santacatterina, M., Lim, S.-N., and Zabih, R.
\newblock Stable estimation of survival causal effects.
\newblock \emph{arXiv preprint arXiv:2310.02278}, 2023.

\bibitem[Ren et~al.(2020)Ren, Yu, Ma, Zhao, Yi, et~al.]{balanced-meta-softmax}
Ren, J., Yu, C., Ma, X., Zhao, H., Yi, S., et~al.
\newblock Balanced meta-softmax for long-tailed visual recognition.
\newblock \emph{Advances in neural information processing systems}, 33:\penalty0 4175--4186, 2020.

\bibitem[Rubin(1976)]{rubin-missingdata}
Rubin, D.~B.
\newblock Inference and missing data.
\newblock \emph{Biometrika}, 63\penalty0 (3):\penalty0 581--592, 1976.

\bibitem[Saerens et~al.(2002)Saerens, Latinne, and Decaestecker]{mlls}
Saerens, M., Latinne, P., and Decaestecker, C.
\newblock Adjusting the outputs of a classifier to new a priori probabilities: a simple procedure.
\newblock \emph{Neural computation}, 14\penalty0 (1):\penalty0 21--41, 2002.

\bibitem[Schmutz et~al.(2022)Schmutz, Humbert, and Mattei]{schmutz-drloss}
Schmutz, H., Humbert, O., and Mattei, P.-A.
\newblock Don't fear the unlabelled: safe semi-supervised learning via simple debiasing.
\newblock \emph{arXiv preprint arXiv:2203.07512}, 2022.

\bibitem[Shi et~al.(2019)Shi, Blei, and Veitch]{dragonnet}
Shi, C., Blei, D., and Veitch, V.
\newblock Adapting neural networks for the estimation of treatment effects.
\newblock \emph{Advances in neural information processing systems}, 32, 2019.

\bibitem[Sohn et~al.(2020)Sohn, Berthelot, Carlini, Zhang, Zhang, Raffel, Cubuk, Kurakin, and Li]{fixmatch}
Sohn, K., Berthelot, D., Carlini, N., Zhang, Z., Zhang, H., Raffel, C.~A., Cubuk, E.~D., Kurakin, A., and Li, C.-L.
\newblock Fixmatch: Simplifying semi-supervised learning with consistency and confidence.
\newblock \emph{Advances in neural information processing systems}, 33:\penalty0 596--608, 2020.

\bibitem[Sportisse et~al.(2023)Sportisse, Schmutz, Humbert, Bouveyron, and Mattei]{arelabelsinformative}
Sportisse, A., Schmutz, H., Humbert, O., Bouveyron, C., and Mattei, P.-A.
\newblock Are labels informative in semi-supervised learning? estimating and leveraging the missing-data mechanism.
\newblock In \emph{International Conference on Machine Learning}, pp.\  32521--32539. PMLR, 2023.

\bibitem[Sun et~al.(2023)Sun, Murphy, Ebrahimi, and D'Amour]{ttsla}
Sun, Q., Murphy, K.~P., Ebrahimi, S., and D'Amour, A.
\newblock Beyond invariance: test-time label-shift adaptation for addressing ``spurious''' correlations.
\newblock \emph{Advances in Neural Information Processing Systems}, 36:\penalty0 23789--23812, 2023.

\bibitem[Tsiatis(2006)]{tsiatis-missingdata}
Tsiatis, A.~A.
\newblock \emph{Semiparametric theory and missing data}, volume~4.
\newblock Springer, 2006.

\bibitem[Van~der Vaart(2000)]{asymptoticstatistics}
Van~der Vaart, A.~W.
\newblock \emph{Asymptotic statistics}, volume~3.
\newblock Cambridge university press, 2000.

\bibitem[Wager \& Athey(2018)Wager and Athey]{wager2018estimation}
Wager, S. and Athey, S.
\newblock Estimation and inference of heterogeneous treatment effects using random forests.
\newblock \emph{Journal of the American Statistical Association}, 113\penalty0 (523):\penalty0 1228--1242, 2018.

\bibitem[Wei et~al.(2021)Wei, Sohn, Mellina, Yuille, and Yang]{crest}
Wei, C., Sohn, K., Mellina, C., Yuille, A., and Yang, F.
\newblock Crest: A class-rebalancing self-training framework for imbalanced semi-supervised learning.
\newblock In \emph{Proceedings of the IEEE/CVF conference on computer vision and pattern recognition}, pp.\  10857--10866, 2021.

\bibitem[Wei \& Gan(2023)Wei and Gan]{acr}
Wei, T. and Gan, K.
\newblock Towards realistic long-tailed semi-supervised learning: Consistency is all you need.
\newblock In \emph{Proceedings of the IEEE/CVF Conference on Computer Vision and Pattern Recognition}, pp.\  3469--3478, 2023.

\bibitem[Wei et~al.(2024)Wei, Mao, Zhou, Wan, and Zhang]{lsc}
Wei, T., Mao, Z., Zhou, Z.-H., Wan, Y., and Zhang, M.-L.
\newblock Learning label shift correction for test-agnostic long-tailed recognition.
\newblock In \emph{Forty-first International Conference on Machine Learning}, 2024.

\bibitem[Xie et~al.(2020)Xie, Luong, Hovy, and Le]{google-selftraining}
Xie, Q., Luong, M.-T., Hovy, E., and Le, Q.~V.
\newblock Self-training with noisy student improves imagenet classification.
\newblock In \emph{Proceedings of the IEEE/CVF conference on computer vision and pattern recognition}, pp.\  10687--10698, 2020.

\bibitem[Zhang et~al.(2023)Zhang, Jennings, Hilmkil, Pawlowski, Zhang, and Ma]{zhang2023towards}
Zhang, J., Jennings, J., Hilmkil, A., Pawlowski, N., Zhang, C., and Ma, C.
\newblock Towards causal foundation model: on duality between causal inference and attention.
\newblock \emph{arXiv preprint arXiv:2310.00809}, 2023.

\bibitem[Zhao et~al.(2022)Zhao, Zhou, Duan, Wang, Qi, and Shi]{dcssl}
Zhao, Z., Zhou, L., Duan, Y., Wang, L., Qi, L., and Shi, Y.
\newblock Dc-ssl: Addressing mismatched class distribution in semi-supervised learning.
\newblock In \emph{Proceedings of the IEEE/CVF conference on computer vision and pattern recognition}, pp.\  9757--9765, 2022.

\bibitem[Zhu et~al.(2024)Zhu, Ding, Jacobson, Wu, Zhan, Jordan, and Jiao]{drst}
Zhu, B., Ding, M., Jacobson, P., Wu, M., Zhan, W., Jordan, M., and Jiao, J.
\newblock Doubly-robust self-training.
\newblock \emph{Advances in Neural Information Processing Systems}, 36, 2024.

\end{thebibliography}
\bibliographystyle{icml2025}

%%%%%%%%%%%%%%%%%%%%%%%%%%%%%%%%%%%%%%%%%%%%%%%%%%%%%%%%%%%%%%%%%%%%%%%%%%%%%%%
%%%%%%%%%%%%%%%%%%%%%%%%%%%%%%%%%%%%%%%%%%%%%%%%%%%%%%%%%%%%%%%%%%%%%%%%%%%%%%%
% APPENDIX
%%%%%%%%%%%%%%%%%%%%%%%%%%%%%%%%%%%%%%%%%%%%%%%%%%%%%%%%%%%%%%%%%%%%%%%%%%%%%%%
%%%%%%%%%%%%%%%%%%%%%%%%%%%%%%%%%%%%%%%%%%%%%%%%%%%%%%%%%%%%%%%%%%%%%%%%%%%%%%%
\newpage
\appendix
\onecolumn

\clearpage
% \setcounter{page}{1}
% \maketitlesupplementary
\begin{center}
Supplementary Material
\end{center}

% {
%     \onecolumn
%     \centering
%     \Large
%     \textbf{\thetitle}\\
%     \vspace{0.5em}Supplementary Material \\
%     \vspace{1.0em}
% }

\section{Proof of \cref{theorem:dr}}
We require some additional regularity assumptions:
\begin{assumption} 1) The number of classes $C$ is bounded w.r.t the number of samples $N$, 2) the missingness mechanism $P(A=1|Y,\theta)$, as well as its estimated counterpart $P(A=1|Y,\theta)$, are bounded below by some constant $\epsilon > 0$, 3) the quantities $P(Y|X,\theta)$ and $P(A|Y,\theta)$ are estimated using auxiliary samples independent of samples used for the sample averaging.
\label{assumption:extra}
\end{assumption}
Assumptions 1 and 2 are natural. For the missingness mechanism, the ground truth being bounded means that there is a non-vanishing proportion of samples for every class. The boundedness of the estimate can be enforced by clipping the estimate. Assumption 3 is called sample splitting in \cite{kennedy-dr}.

For convenience we use operator $\E_N$ to denote the average of $N$ samples i.e. $\frac{1}{N}\sum_{i=1}^N$. Note that this is by itself a random variable, in contrast to $\E$ which is a fixed number.

\begin{proof}[Proof of \cref{theorem:dr}] Because $C$ is bounded (assumption \ref{assumption:extra}), we can fix a class $c$ and prove the theorem.
Let us define the influence function $\phi$, parameterized by $\theta$, as
\begin{equation}
\phi(O | \theta)(c) = P(Y=c|X,\theta) + \frac{\one(A=1)}{P(A=1|Y,\theta)} (\one(Y=c) - P(Y=c|X,\theta)) - P(Y=c)
\end{equation}
As we have done in the main text, we use $\phi(O)$ to denote the same function but all estimated quantities are replaced with their truths. In other words, we use $\phi(O)$ for $\phi(O|\theta_0)$ where $\theta_0$ is the truth, given that our model contains $\theta_0$ e.g. when the model is consistent.

Recall that:
\begin{equation}
\begin{aligned}
\Psi_{dr}(\theta)(c) &= \frac{1}{N}\sum_{i=1}^N \left\{P(Y=c|X,\theta) + \frac{\one(A=1)}{P(A=1|Y,\theta)} (\one(Y=c) - P(Y=c|X,\theta))\right\}\\
&= \E_N [\phi(O|\theta)(c)] + P(Y=c)
\end{aligned}
\end{equation}

We will show that:
\begin{equation}
\Psi_{dr}(\theta)(c) - P(Y=c) = (\E_N - \E)[\phi(O)(c)] + o_P(N^{-1/2})
\label{eq:proof-linearity}
\end{equation}
To do that, we use the following decomposition
\begin{equation}
\begin{aligned}
\Psi_{dr}(\theta)(c) - P(Y=c) &= \E_N [\phi(O|\theta)(c)] \\
&= (\E_N - \E)[\phi(O)(c)] + (\E_N - \E)[\phi(O|\theta)(c) - \phi(O)(c)] + \E[\phi(O|\theta)(c)]
% &+ (\E_n - \E)[\phi(O;\theta) - \phi(O)]\\
% &+ \E[P(Y=c|X,\theta)] - \E[P(Y=c|X)] + \E[\phi(O,\theta)]
\end{aligned}
\end{equation}
and analyze the second and third term. The third term is:
\begin{equation}
\begin{aligned}
\E[\phi(O|\theta)(c)] &= \E[P(Y=c|X,\theta)] + \E\left[\frac{\one(A=1)}{P(A=1|Y,\theta)}(\one(Y=c) - P(Y=c|X,\theta))\right]- P(Y=c) \\
&= \E\left[P(Y=c|X,\theta) + \frac{P(A=1|Y)}{P(A=1|Y,\theta)}(P(Y=c|X) - P(Y=c|X,\theta))\right] - \E[P(Y=c|X)]\\
&= \E\left[(P(Y=c|X,\theta) - P(Y=c|X)) (P(A=1|Y,\theta) -P(A=1|Y)) \frac{1}{P(A=1|Y,\theta)}\right]\\
\end{aligned}
\end{equation}
by Cauchy-Schwarz inequality:
\begin{equation}
\begin{aligned}
\E[\phi(O|\theta)(c)] &\le \frac{1}{\epsilon} \|P(A=1|Y,\theta) - P(A=1|Y)\|_2 \|P(Y=c|X,\theta) - P(Y=c|X)\|_{L_2(P)}\\
&= \frac{1}{\epsilon} o_P(N^{-1/4} N^{-1/4}) = o_P(N^{-1/2})
\end{aligned}
\end{equation}
by assumption \ref{assumption:4th-root-n} and that $P(A=1|Y,\theta) > \epsilon$ (assumption \ref{assumption:extra}). The second term can be bounded by Chebyshev inequality
% \begin{equation}
% \begin{aligned}
% \E[\E_N[\phi(O|\theta)(c) - \phi(O)(c)]] &= \E[\phi(O|\theta)(c) - \phi(O)(c)]\\
% \var[\E_N[\phi(O|\theta)(c) - \phi(O)(c)]] &= \frac{1}{N}\var[\phi(O|\theta)(c) - \phi(O)(c)] \le 
% \end{aligned}
% \end{equation}
\begin{equation}
P(|(\E_N - \E)[\phi(O|\theta)(c) - \phi(O)(c)]| \ge t) \le \frac{\var[\E_N[\phi(O|\theta)(c) - \phi(O)(c)]]}{t^2} = \frac{\var[\phi(O|\theta)(c) - \phi(O)(c)]}{Nt^2}
\end{equation}
note here that $\theta$ is independent of the samples used for $\E_N$ by assumption \ref{assumption:extra}. For any $\varepsilon > 0$, by picking $t = \frac{1}{\sqrt{N\varepsilon}}$ we get
\begin{equation}
P\left(\left|\frac{(\E_N - \E)[\phi(O|\theta)(c) - \phi(O)(c)]}{N^{-1/2}}\right| \ge \frac{1}{\sqrt{\varepsilon}}\right) \le \varepsilon \var[\phi(O|\theta)(c) - \phi(O)(c)]
\end{equation}
by the definition of $O_P$, we then get
\begin{equation}
(\E_N - \E)[\phi(O|\theta)(c) - \phi(O)(c)] = O_P(N^{-1/2}\var[\phi(O|\theta)(c) - \phi(O)(c)])
\end{equation}
Because $\phi$ is a continuous function of $P(Y|X,\theta)$ and $P(A|Y,\theta)$ (given $P(A|Y,\theta) > \epsilon$, assumption \ref{assumption:extra}), by the continuous mapping theorem and the fact that $P(Y|X,\theta)$ and $P(A|Y,\theta)$ are convergent in probability (assumption \ref{assumption:4th-root-n}), we get $\var[\phi(O|\theta)(c) - \phi(O)(c)] = o_P(1)$. This gives
\begin{equation}
(\E_N - \E)[\phi(O|\theta)(c) - \phi(O)(c)] = o_P(N^{-1/2})
\end{equation}
Therefore, we have shown that the second and third term are both $o_P(N^{-1/2})$, proving \cref{eq:proof-linearity}. As the final step, multiply both sides of this equation by $\sqrt{N}$ we get:
\begin{equation}
\sqrt{N}(\Psi_{dr}(\theta)(c) - P(Y=c)) = \sqrt{N} (\E_N - \E)[\phi(O)(c)] + o_P(1) \rightsquigarrow \mathcal{N}(0, \var[\phi(O)(c)])
\end{equation}
by the central limit theorem, and $\var[\phi(O)(c)] = \E[\phi(O)(c)^2]$ because $\E[\phi(O)(c)] = 0$.
\end{proof}

While we started with the definition of $\phi$, \cref{eq:proof-linearity} shows that $\phi$ is indeed an influence function. Now we show that $\phi$ is also the efficient influence function, by using the characterization of the model's tangent space \cite{tsiatis-missingdata}. Note that the joint probability factorizes as $P(X,A,Y) = P(X)P(Y|X)P(A|Y)$, therefore the tangent space $\mathcal{T}$ factorizes as $\mathcal{T} = \mathcal{T}_{X} \oplus \mathcal{T}_{Y|X} \oplus \mathcal{T}_{A|Y}$ where $\mathcal{T}_X = \{h(X): \E[h] = 0\}$, $\mathcal{T}_{Y|X} = \{h(X,Y): \E[h|X] = 0\}$, $\mathcal{T}_{A|Y} = \{h(A,Y): \E[h|Y] = 0\}$, and the 3 subspaces are pairwise orthogonal. All influence functions are orthogonal to the tangent space, but the influence function that is also in the tangent space has the smallest variance and is called the efficient influence function. As $\phi$ is already an influence function, we need only show that $\phi$ is in $\mathcal{T}$. We write $\phi$ as
\begin{equation}
\phi(O)(c) = (P(Y=c|X) - P(Y=c)) + \left[\frac{\one(A=1)}{P(A=1|Y)} - 1\right](\one(Y=c) - P(Y=c|X)) + (\one(Y=c) - P(Y=c|X))
\end{equation}
and note that the first, second and third term are in $\mathcal{T}_X$, $\mathcal{T}_{A|Y}$ and $\mathcal{T}_{Y|X}$ respectively. Therefore, $\phi$ is indeed in $\mathcal{T}$. The efficient influence function has the smallest variance of all influence function, and therefore our estimator being asymptotically linear in $\phi$ (\cref{eq:proof-linearity}) has the smallest mean squared error in a local asymptotic minimax sense \cite{kennedy-dr, asymptoticstatistics}

\section{Further background and related work}
\paragraph{Discussion on semi-supervised EM.}
It appears that semi-supervised EM was first used for parameter estimation when the missingness mechanism is non-ignorable in \cite{ibrahim1996parameter}, but has not been used for label shift estimation.
Perhaps this is because the semi-supervised situation where additional unlabeled data is available during training is rarer than the test-time adaptation case. EM is well suited to take advantage of the extra unlabeled data to improve the classifier under very scarce and long-tailed labeled data. While the connection between pseudo-labeling and EM has been explored before \cite{entropyminimization}, the situation with label shift has not until recently \cite{simpro}. Here the application of EM is much more interesting, because other than simply giving pseudo-labeling a rigorous formulation, EM also estimates the missingness mechanism (equivalently the label distribution shift), which is important for shift correction and thus high-quality pseudo-labels \cite{acr}. The application of confidence thresholding can be seen as a sparse variant of EM \cite{neal1998view}.

\paragraph{The doubly-robust risk.} 
\label{subsec:dr-risk}
A technique that also derives from the theory of semi-parametric efficiency is orthogonal statistical learning \citep{foster2023orthogonal}. The idea is to minimize the doubly-robust risk:
\label{subsec:method-dr-risk}
\begin{equation}
\label{eq:dr-risk}
\mathcal{R}(\theta_2) = \frac{1}{N} \sum_{i=1}^N \Bigg[ l(x_i, \hat y_i|\theta_2) + \frac{\one(a_i=1)}{P(A=a_i|Y=y_i, \theta_1)} (l(x_i, y_i | \theta_2) - l(x_i, \hat y_i | \theta_2))\Bigg]
\end{equation}
where $l(x,y|\theta) = -\sum_{c=1}^C [y]_c \log P(Y=c|X=x,\theta)$ is the negative cross-entropy. 
The notation $[y]_c$ means that we are using the $c$-entry in a C-dimension probability vector $y$. 
Thus, $y_i$ denotes the one-hot label of observation $i$, while $\hat y_i$ denotes the pseudo-label, which can be one-hot or all-zero. 
Finally, we use $\theta_1$ to denote that $P(a|y,\theta_1)$ is an estimation from a previous stage, but it can be estimated with $\theta_2$ as well. 
The risk $\mathcal{R}(\theta_2)$ can be used as a training loss in a straightforward fashion. 
Similar to the doubly robust estimation of $P(Y)$, the doubly robust risk provides approximately unbiased estimation of the risk. 
This property has been used in \citep{arelabelsinformative, onnonrandommissinglabels, drst} also in the semi-supervised learning setting.
More broadly, it is at the heart of one of the core techniques in heterogenous treatment effect estimation in causal estimation \cite{kennedy2023towards, foster2023orthogonal, wager2018estimation}. 
The focus here is not the estimation of $\mathcal{R}(\theta_2)$ per se, but the quality of the learned model \cite{foster2023orthogonal}.
By using the doubly-robust risk, we can achieve an optimality result similar in spirit to our theorem \cref{theorem:dr}, but for the generalization error.
While this is appealing, in practice there are 2 problems with this approach. First, the inverse probability weight $P(A=a_i|Y=y_i,\theta_1)$ can be very large if the class ratio is highly unlabeled, making training unstable \cite{kallus2020deepmatch, pham2023stable}. 
This problem exists for our estimation as well. However, it is much easier to control for estimation than for training because of the iterative nature of model update. Secondly, we can further write $\mathcal{R}$ as:
\begin{equation}
\mathcal{R}(\theta_2) = \frac{1}{N}\sum_{i=1}^N l\left(x_i, \hat y_i + \frac{\one(a_i=1)}{P(A=a_i|Y=y_i,\theta_1)} (y_i - \hat y_i)\Bigg\vert\theta_2\right)
\end{equation}
which is a cross-entropy loss with new meta-pseudo-labels. However, these labels are not meant to be learned exactly, and furthermore they can be negative. Thus, theoretical works have to put stringent assumptions on the models. In \cref{subsec:ablation-1}, we show that experimentally that the instability problem makes doubly-robust risk performance worse than our 2-stage approach.

\section{Training and hyperparameter settings.}
\label{subsec:training-setting}
For neural network training, we follow the implementation and hyperparameter settings of \cite{simpro}. In particular, we adapt the core code of SimPro for Supervised, MLE and EM. For MLE, we update $P(A|Y)$ using the Adam optimizer with learning rate 1e-3, while for EM we use a momentum update similar to SimPro's update of $P(Y|A)$ because it has a a closed-form solution at each mini-batch. We use Wide ResNet-28-2 on all methods and all datasets in this section, including Imagenet-127, because we are motivated by the fact that stage-1's goal is not classification accuracy but the estimation of a finite-dimensional parameter. When using Wide ResNet-28-2 for Imagenet-127, we use the hyperparameters of CIFAR-100, except we lower the batch size of unlabeled data to 2 times that of labeled data instead of 8 for memory reason. We do not perform additional hyperparameter tuning. All experiments can be performed on 1 A6000 RTX GPU, and are run 3 times. We report the total variation distance between the estimated and the ground truth unlabeled class distribution, similar to its usage in Theorem 3.1 of \cite{lsc}, and the top-1 classification accuracy.

In the second stage of our algorithm, we freeze our estimation and plug it in SimPro and BOAT.
We keep exactly the same hyperparameter settings that SimPro and BOAT use. In particular, for Imagenet-127, we now use ResNet-50 and run each experiment once.
In SimPro, we set the unlabeled class distribution $P(Y|A=0)$ at the E-step;  however, we still keep a running estimate of the class distribution $P(Y)$ in the logit adjustment loss \cref{eq:simpro-la-loss}. While it is possible to use the first stage estimate in the logit adjustment loss, we observe that doing so results in lower accuracy than using the the running average. This is conceptually consistent with the role of the running average - serving not as an accurate estimate of $P(Y)$ but to make the classifier's class distribution uniform through the logit adjustment loss, which is good for the test set. Similarly, in BOAT, we only replace $\Delta_c = \log P(Y|A=1) - \log P(Y|A=0)$ in equation (4) of \cite{boat}, which is adjusting a classifier's predictions from the labeled to the unlabeled class distribution, with our SimPro + DR estimate instead of their on-the-fly estimate.

% \section{Additional experiments}
% % \input{tables/cifar10-tv}
% \input{tables/cifar100-tv}

%%%%%%%%%%%%%%%%%%%%%%%%%%%%%%%%%%%%%%%%%%%%%%%%%%%%%%%%%%%%%%%%%%%%%%%%%%%%%%%
%%%%%%%%%%%%%%%%%%%%%%%%%%%%%%%%%%%%%%%%%%%%%%%%%%%%%%%%%%%%%%%%%%%%%%%%%%%%%%%

\end{document}